%% file: main.tex
\documentclass[final,5p,times,twocolumn,authoryear]{elsarticle}
\input{preamble}
\usepackage{stfloats}
\begin{document}
\begin{frontmatter}
\title{Filling Before Advancing: Capability-Gap-Driven Post-Training for
Scenario-Specialized Remote Sensing MLLMs}
\author[aff1]{Yuheng Zong}
\author[aff1]{Minghua Wang\corref{cor1}}
\ead{wangminghua@nankai.edu.cn}
\author[aff1]{Xin Zhao}
\author[aff2]{Zhi-Hui Zhan}
\author[aff3]{Antonio Plaza}
\author[aff4]{J{\'o}n Atli Benediktsson}
\cortext[cor1]{Corresponding author}
\affiliation[aff1]{organization={The Institute of Robotics and Automatic Information System (IRAIS),
the Tianjin Key Laboratory of Intelligent Robotics (tjKLIR), Nankai University},
city={Tianjin}, postcode={300071}, country={China}}
\affiliation[aff2]{organization={The College of Artificial Intelligence, Nankai University},
city={Tianjin}, postcode={300071}, country={China}}
\affiliation[aff3]{organization={Hyperspectral Computing Laboratory,
Department of Technology of Computers and Communications,
Escuela Polit\'{e}cnica, University of Extremadura}, city={C\'{a}ceres}, country={Spain}}
\affiliation[aff4]{organization={Faculty of Electrical and Computer Engineering,
University of Iceland}, city={Reykjav\'{i}k}, country={Iceland}}
\begin{abstract}
Remote sensing multimodal large language models (RS-MLLMs) have improved general aerial-image understanding. However, Earth observation applications require fine-grained scenario specialization, constrained by scarce high-quality scenario data and incomplete capability coverage. We formulate this adaptation as a capability-gap-driven post-training problem and propose \textit{filling before advancing} (FBA). Rather than relying on single-stage supervised fine-tuning (SFT) over target-domain samples, FBA first fills prerequisite capability gaps before advancing toward scenario specialization. We instantiate FBA for coastal harbor understanding, a multi-source scenario with individual RGB, SAR, PAN, and NIR images, by constructing CPRS (Coastal-Port Remote Sensing), a three-layer supervision dataset coupled with three ordered stages: (1) RS semantic anchoring for overhead-view visual-language alignment, (2) domain-bridge convergence for shared RS priors across target and bridging scenarios under different modalities, and (3) evidence-grounded scenario tuning for downstream performance. We construct HarborEval, an eight-track diagnostic benchmark covering perception, spatial understanding, robustness, and generation. With the same task-specific supervision pool, the HarborEval score increases from 57.95 with Direct-SFT to 70.29 with FBA on LLaVA-v1.5, and from 81.09 to 83.37 on Qwen3-VL. FBA also outperforms Collapsed-SFT, which combines bridging and scenario supervision in one stage, and leads on harbor-related VRSBench/RSVQA subsets and OpenEval. Stage-wise and role-replacement analyses support the distinct supervision roles of the stages. A lightweight urban instantiation further demonstrates the cross-scenario applicability of FBA, with the proposed stage order yielding stronger overall performance than the tested alternatives. \textbf{Public examples and release updates for
CPRS, HarborEval, code, and trained weights are available at}
\href{https://github.com/Z0ngL1ng/filling-before-advancing}{%
    \renewcommand{\UrlFont}{\bfseries\itshape}%
    \nolinkurl{https://github.com/Z0ngL1ng/filling-before-advancing}%
}.
\end{abstract}

\end{frontmatter}

\section{Introduction}

Multimodal large language models (MLLMs) have expanded visual understanding from closed-set recognition to open-world perception, logical reasoning, and instruction following~\citep{radford2021learning,alayrac2022flamingo,li2023blip2,dai2023instructblip,liu2023visual,bai2025qwen25vl}. This development has also extended to remote sensing (RS), enabling RS-MLLMs capable of aerial image captioning, visual question answering, region-level instruction following, and grounded spatial interpretation~\citep{hu2025rsgpt,kuckreja2024geochat,zhang2024earthgpt,muhtar2024lhrsbot,pang2025vhm,soni2025earthdial,zhan2025skyeyegpt}. Instead of producing category labels or boxes, RS-MLLMs connect visual evidence with natural-language queries, raising the question of how these general capabilities can be adapted to the requirements of specific RS scenarios.

\begin{figure*}[!t]
\centering
\includegraphics[width=\textwidth,keepaspectratio]{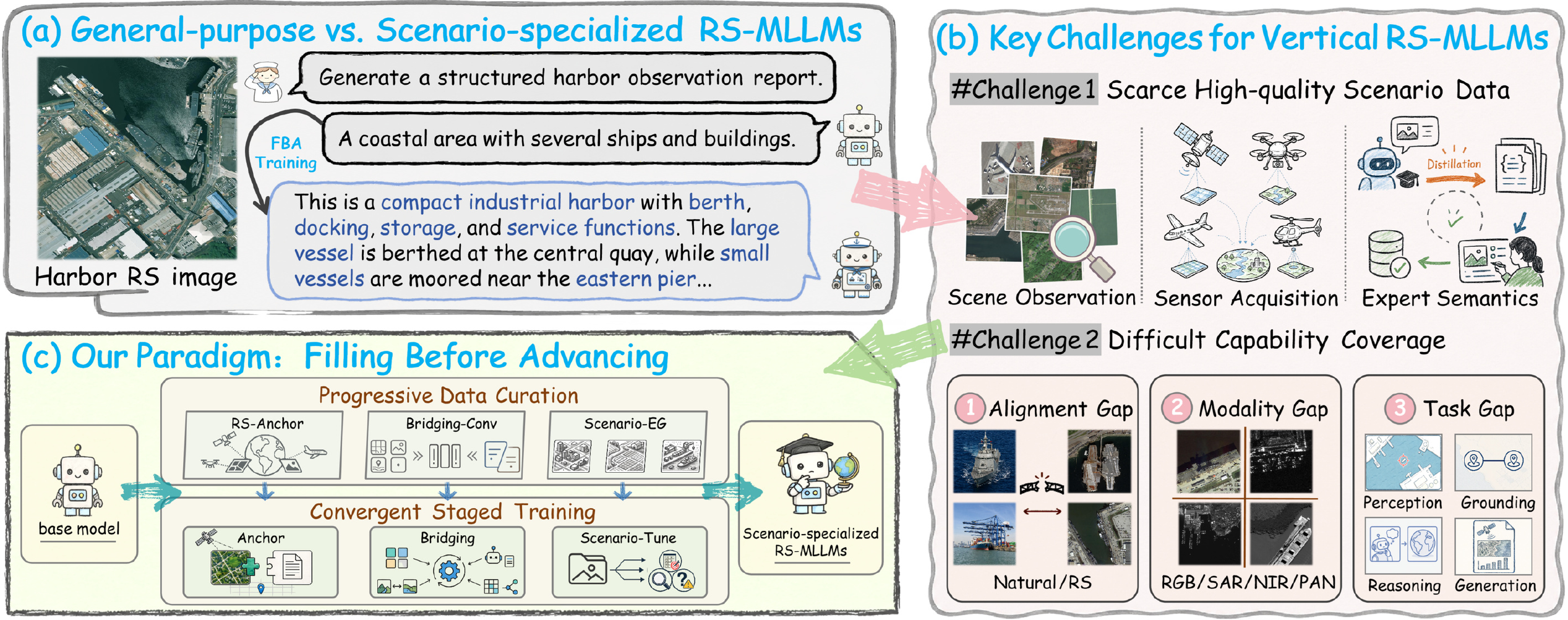}
\caption{Motivation and paradigm of the proposed FBA.}
\label{fig:motivation_overview}
\end{figure*}

Real-world RS applications rely on the interpretation and understanding of targeted task scenarios, rather than broad scene categories~\citep{li2024visionlanguage}. Taking coastal harbor monitoring as a representative case, RS-MLLMs are expected to surpass simple and generic descriptions, such as water, roads, buildings, and ships. As illustrated in Fig.~\ref{fig:motivation_overview}(a), general RS-MLLMs may describe what exists without organizing the evidence for the scenario. Scenario-specialized RS-MLLMs are required to generate a structured and evidence-grounded report of operationally meaningful harbor attributes, including functional zones, vessel scale, and spatial layout. This discrepancy reveals a gap between universal capability and scenario-specialized usability of RS-MLLMs.

A straightforward solution is to collect harbor instructions and directly fine-tune a general RS-MLLM, yet this is often insufficient due to two challenges, shown in Fig.~\ref{fig:motivation_overview}(b). First, high-quality, task-matched harbor supervision is scarce. General MLLMs typically rely on massive natural image--text corpora for training~\citep{alayrac2022flamingo}. In contrast to readily accessible natural images, expert-annotated supervision tailored to harbor tasks remains scarce, because long-term scene observation, multi-sensor acquisition, and rigorous expert annotation are time-consuming and labor-intensive~\citep{wang2024skyscript,kuckreja2024geochat}. Second, achieving comprehensive capability coverage for RS specialization is difficult. When migrating from general MLLMs to RS-MLLMs, gaps in alignment, modality, and task arise for specific RS scenarios, while single-stage training paradigms, typified by direct supervised fine-tuning (SFT), may not fully bridge these gaps. These gaps result from the shift in viewing angles, differences among RS sensors, and specialized task requirements from natural to RS scenes~\citep{liu2024remoteclip,zhang2024earthgpt,muhtar2024lhrsbot,soni2025earthdial}. With direct SFT, these limited task-matched samples must therefore support both prerequisite adaptation to RS imagery and final harbor-task learning.

This motivates the \textit{filling before advancing} (FBA) perspective of scenario-specialized RS-MLLM post-training, displayed in Fig.~\ref{fig:motivation_overview}(c). FBA assigns broader RS supervision to prerequisite capability development and concentrates task-matched harbor supervision on final specialization. Here, \textit{filling} develops prerequisite capabilities in overhead-view semantics and sensor-aware interpretation through RS anchoring and multi-source bridging. \textit{Advancing} then focuses adaptation on harbor-specific tasks, including evidence-grounded responses and rejection. In this capability-gap-driven design, the identified gaps guide the supervision content and the role of each training stage.

In this paper, we instantiate the proposed paradigm in the context of coastal harbor understanding. As illustrated in Fig.~\ref{fig:data_curation_pipeline}, we establish CPRS, a three-layer supervision dataset comprising RS-Anchor, Bridging-Conv, and Scenario-EG in alignment with the three ordered adaptation stages. RS semantic anchoring establishes broad overhead-view visual-language alignment as the basis for subsequent specialization. Domain-bridge convergence learns RS priors shared across target and bridging scenarios from RGB, SAR, PAN, and NIR samples. Each sample uses an individual image from its corresponding modality, allowing supervision across sensors without requiring co-registered multi-sensor inputs. Evidence-grounded scenario tuning focuses adaptation on harbor-specific downstream tasks and strengthens evidence-grounded responses under negative or ambiguous conditions.
An eight-track diagnostic benchmark is constructed and denoted as HarborEval, covering RGB, SAR, PAN, and NIR imagery from both harbor and non-harbor scenes. Using HarborEval, we compare FBA with direct tuning and collapsed training on two representative backbones. Collapsed training combines bridging and scenario supervision in a single stage. FBA achieves higher overall HarborEval scores than Direct-SFT and Collapsed-SFT with both LLaVA-v1.5 and Qwen3-VL backbones~\citep{liu2024improved,bai2025qwen3vl}. Further comparisons with existing RS-MLLMs on HarborEval, the harbor-related subsets of VRSBench and RSVQA~\citep{li2024vrsbench,lobry2020rsvqa}, and OpenEval support the effectiveness of the proposed route. Stage-wise analyses and role-replacement controls further examine the intended capability role of each stage along the ordered route. A lightweight urban instantiation on Qwen3-VL-2B-Instruct further demonstrates the cross-scenario applicability of FBA, with route and stage-order comparisons favoring semantic anchoring and bridging before final scenario tuning.
The main contributions of this work are as follows:
\begin{itemize}
    \item We formulate scenario-specialized RS-MLLM adaptation under scarce high-quality, task-matched scenario supervision as a capability-gap-driven post-training problem.
    \item We propose FBA, a post-training route that progressively fills capability gaps through three ordered stages, including RS semantic anchoring, domain-bridge convergence, and evidence-grounded scenario tuning, respectively supported by the CPRS supervision layers RS-Anchor, Bridging-Conv, and Scenario-EG.
    \item We instantiate FBA for coastal harbor understanding and construct HarborEval, an eight-track diagnostic benchmark. Extensive experiments show that FBA consistently outperforms alternative post-training routes across multiple backbones and achieves competitive performance against existing RS-MLLMs, with stage-wise analyses supporting the distinct roles of the ordered adaptation stages.
\end{itemize}

\section{Related Work}
\subsection{Remote Sensing Vision-Language Foundations}
RS vision-language research has narrowed the semantic gap between natural-image pretraining and RS imagery through captioning~\citep{lu2018exploring,cheng2022nwpu}, VQA~\citep{lobry2020rsvqa}, cross-modal retrieval~\citep{yuan2022exploring}, and large-scale geospatial image-text alignment~\citep{wang2024skyscript,zhang2024rs5m,liu2024remoteclip}. These works provide broad RS semantic grounding for open-ended understanding.

Sensor differences introduce additional constraints. Spectral pre-training and radar--optical alignment address complementary observation characteristics~\citep{hong2024spectralgpt,fuller2023croma}. Multi-source models extend this adaptation across sensor modalities~\citep{guo2024skysense,soni2025earthdial}.

However, these foundations mainly address broad RS semantic alignment and sensor adaptation and rarely specify how to use scarce scenario-level supervision for modality-aware evidence, spatial grounding, uncertainty handling, and rejection.

\subsection{RS-MLLMs and Scenario-Level Understanding}
Recent RS-MLLMs adapt general multimodal foundations to the RS field through RS alignment, instruction tuning, grounded dialogue, and multi-source Earth-observation inputs~\citep{kuckreja2024geochat,zhang2024earthgpt,soni2025earthdial,zhan2025skyeyegpt}. Such advances support general instruction following, captioning, VQA, and region-level interpretation~\citep{hu2025rsgpt,luo2024skysensegpt,li2025lhrsbotnova}.

Alongside these general capabilities, recent work addresses response reliability. VHM combines factual and deceptive questions for honest instruction following~\citep{pang2025vhm}, complementing visual-grounding and hallucination diagnostics~\citep{li2023evaluating,guan2024hallusionbench}.

Scenario-specific applications require these capabilities to work together under shared evidence constraints. Direct target-domain SFT can overburden limited high-quality scenario data, because the same samples must support both prerequisite adaptation and final scenario behavior.

\subsection{Post-Training Routes for Scenario Specialization}
Our work is also related to instruction tuning~\citep{wei2022finetuned,liu2023visual}, curriculum learning~\citep{bengio2009curriculum}, and continual domain adaptation~\citep{gururangan2020dont,kirkpatrick2017overcoming,rolnick2019experience}. These paradigms show that training order and supervision design affect adaptation, but commonly organize data by example difficulty, instruction diversity, domain continuation, or task arrival. Skill-it! explicitly models prerequisite skill dependencies~\citep{chen2023skillit}, while EarthDial progressively expands multi-source interpretation through staged fine-tuning~\citep{soni2025earthdial}.

FBA builds on these staged adaptation strategies by organizing supervision around the capabilities required for scenario specialization. Each data layer is assigned a distinct role: broad RS semantic anchoring, multi-source domain-bridge convergence, and final evidence-grounded scenario tuning. Unlike D-MoLE's dynamic LoRA-expert allocation and inter-modal curriculum~\citep{ge2025dmole}, FBA uses a fixed supervision-role sequence, examined through stage diagnostics and role-replacement comparisons.

\section{Methods}
FBA progressively develops scenario-specialized capabilities through staged training on three supervision layers: RS-Anchor ($D_1$), Bridging-Conv ($D_2$), and Scenario-EG ($D_3$). These layers support RS semantic anchoring, multi-source bridging, and evidence-grounded scenario tuning. We first describe the construction of these supervision layers in CPRS, then present the staged training procedure and the harbor-scenario evaluation protocol.

\begin{figure*}[!t]
\centering
\includegraphics[width=\textwidth,keepaspectratio]{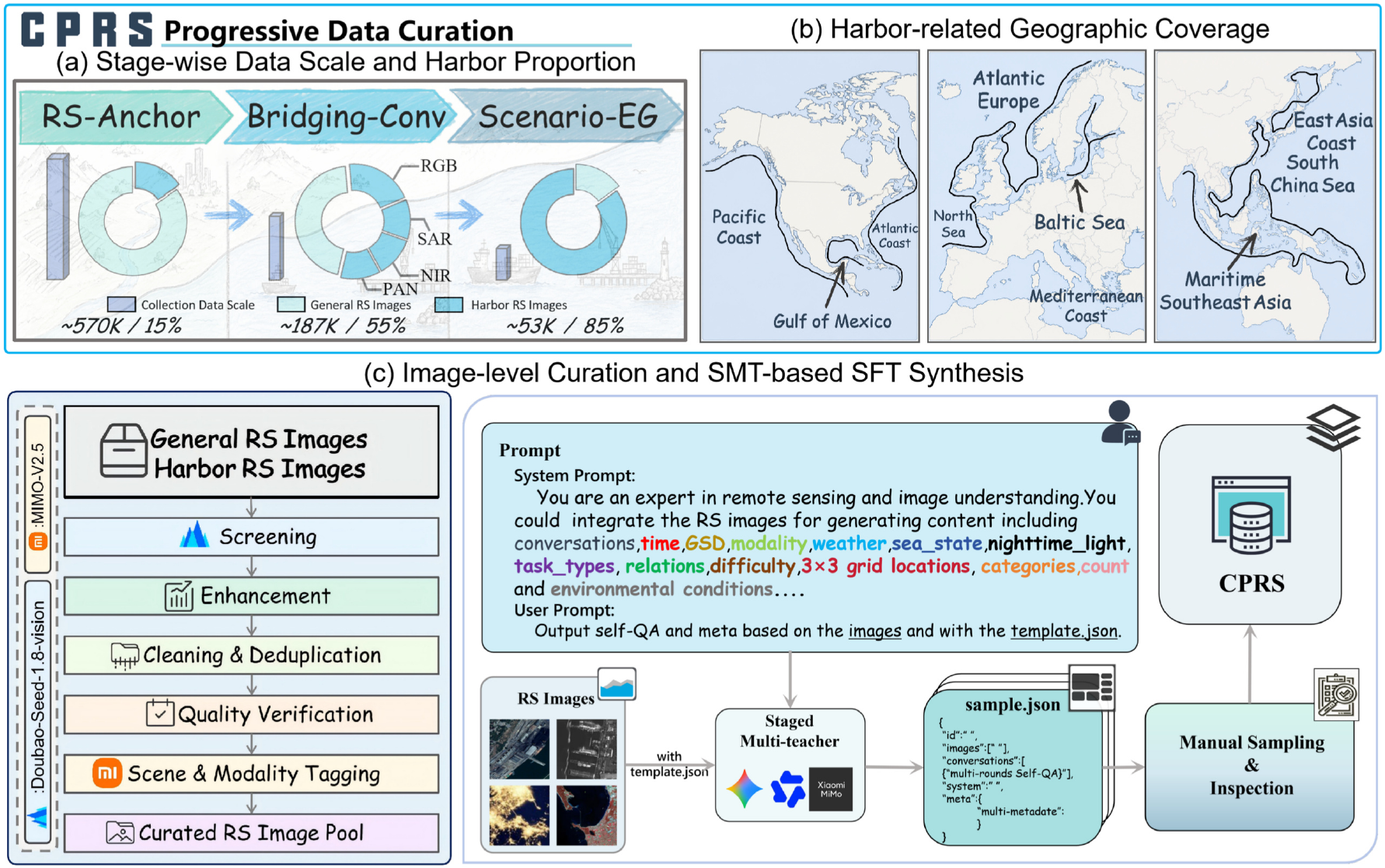}

\caption{Progressive data curation of the CPRS dataset for the staged route.}
\label{fig:data_curation_pipeline}
\end{figure*}

\subsection{Progressive Data Curation}
\label{sec:data_curation}
To demonstrate the practical realization of FBA through coastal harbor understanding, we construct CPRS, a three-layer supervision dataset (Fig.~\ref{fig:data_curation_pipeline} and Table~\ref{tab:appendix_data_source_statistics}). RS-Anchor contains 569,853 RGB image-caption pairs, Bridging-Conv contains 187,296 multi-source SFT records, and Scenario-EG contains 53,000 harbor-specialized instruction records. The harbor-related image proportion increases from 15\% through 55\% to 85\%, while the supervision changes from broad alignment to scenario-grounded responses.

\begin{table}[!htbp]

\centering
\caption{Training-supervision pools used by the progressive route. Evaluation records and answer-bearing fields are excluded from every pool.}
\label{tab:appendix_data_source_statistics}
\small
\setlength{\tabcolsep}{3pt}
\renewcommand{\arraystretch}{1.12}
\begin{tabularx}{\columnwidth}{@{}>{\raggedright\arraybackslash}p{0.37\columnwidth}>{\centering\arraybackslash}Xr@{}}
\toprule
\rowcolor[HTML]{F2F2F2}
\textbf{Pool} & \textbf{Modality} & \textbf{Scale} \\
\midrule
RS-Anchor ($D_1$) & RGB & 569,853 \\
Bridging-Conv ($D_2$) & RGB / SAR / PAN / NIR & 187,296 \\
Scenario-EG ($D_3$) & RGB / SAR / PAN / NIR & 53,000 \\
\bottomrule
\end{tabularx}
\par\vspace{3pt}

\end{table}

The three pools differ in supervision format as well as scene coverage. RS-Anchor retains one caption per image and emphasizes visible land cover, object distribution, and coarse spatial structure. Bridging-Conv converts heterogeneous sources into image-centered instruction records with modality-compatible answers. Scenario-EG concentrates those instructions on harbor-specific decisions and reports. Accordingly, Table~\ref{tab:appendix_data_source_statistics} reports the number of image-caption pairs for $D_1$ and the number of SFT records for $D_2$ and $D_3$.

To align overhead-view visual patterns with RS semantics, RS-Anchor is constructed from existing RGB RS captioning~\citep{lu2018exploring,cheng2022nwpu,ge2025rsteller}, retrieval~\citep{yuan2022exploring}, and geospatial image-text resources~\citep{wang2024skyscript,yuan2025chatearthnet,soni2025earthdial}. These data are screened, normalized, deduplicated, and quality-checked to form a large-scale, quality-controlled image-text supervision pool for initial RS visual-language alignment.

Following the initial RS visual-language alignment established by RS-Anchor, Bridging-Conv extends supervision to target harbor scenes and coastal-port bridging scenes across RGB, SAR, PAN, and NIR. As illustrated in Fig.~\ref{fig:data_curation_pipeline}(c), staged multi-teacher (SMT) distillation constructs $D_2$ through three successive steps. Gemini-3-pro ($T_{\mathrm{meta}}$) extracts image-grounded metadata describing scene content and modality-specific evidence. Qwen3.6-27B ($T_{\mathrm{syn}}$) uses this metadata to generate image-centered instruction--response pairs. MiMo-v2.5 ($T_{\mathrm{ver}}$) reviews the generated pairs for consistency with the image evidence and sensor characteristics, and accepted records form the Bridging-Conv supervision pool. The review accommodates uncertainty in SAR, accounts for the spectral characteristics of NIR, and emphasizes grayscale geometry in PAN~\citep{zhu2021deep,vivone2023multispectral}.

\begin{equation}
\label{eq:bridge_curation}
\begin{aligned}
e_i
&= T_{\mathrm{meta}}(x_i,m_i,s_i),\\
\tilde{u}_i
&= T_{\mathrm{syn}}(x_i,e_i),\\
u_i
&= T_{\mathrm{ver}}(x_i,e_i,\tilde{u}_i),\\
D_2
&= \operatorname{SMT}(\mathcal{X}_2)
 = \left\{
 (x_i,u_i)
 \mid x_i\in\mathcal{X}_2,\;
 u_i\neq\emptyset
 \right\},
\end{aligned}
\end{equation}
Here, $\mathcal{X}_2$ denotes the curated pool of target and bridging images, with $m_i$ and $s_i$ specifying the modality and scene label of each image $x_i$. The variables $e_i$, $\tilde{u}_i$, and $u_i$ denote image-grounded metadata, a candidate instruction--response record, and its verified form, respectively. The value $u_i=\emptyset$ indicates a rejected candidate. The verified records constitute $D_2$.

Scenario-EG constitutes the final supervision layer $D_3$, concentrating supervision on evidence-grounded behavior. It covers harbor-specific downstream tasks and incorporates negative and uncertainty-aware supervision under stricter evidence-grounding criteria. Construction details and teacher prompts are provided in Supplementary Sections~\ref{app:data_curation}--\ref{app:smt_workflow}.

\begin{figure}[!tb]
\centering
\includegraphics[width=0.64\columnwidth]{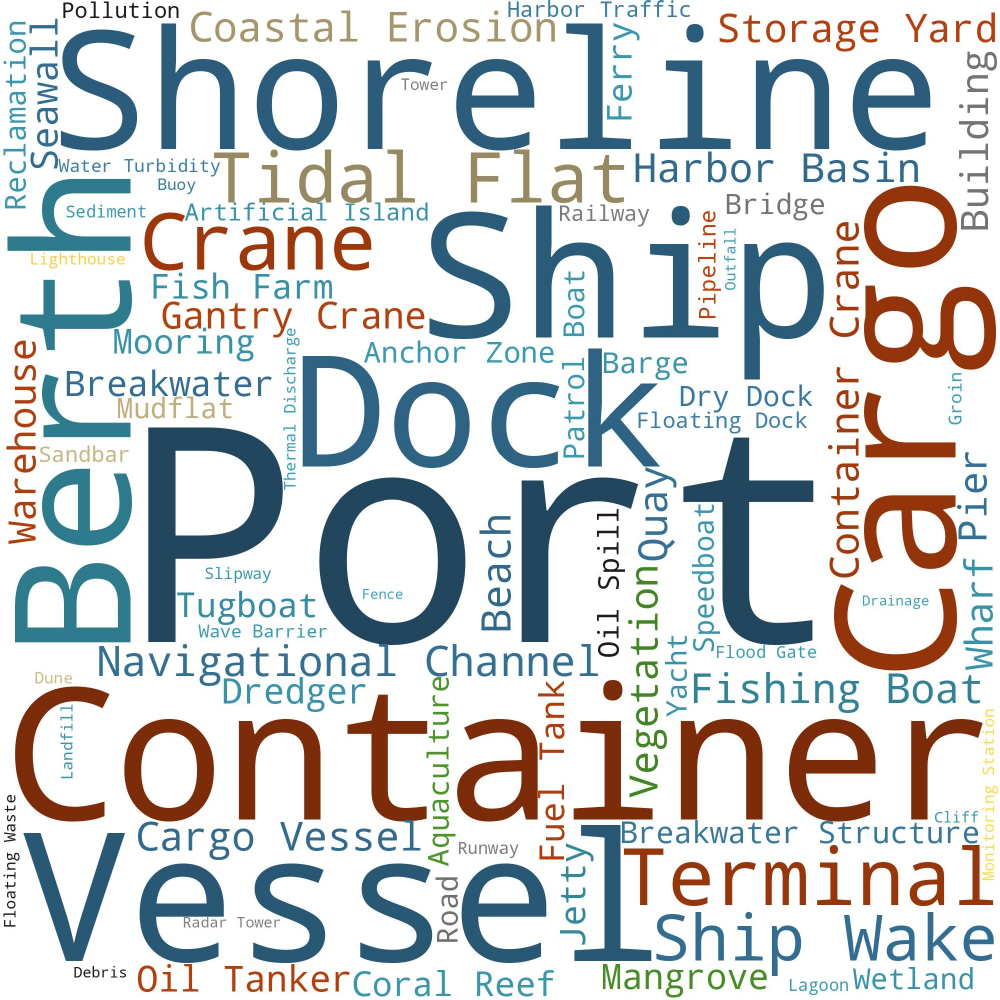}
\caption{Word-cloud summary of harbor-oriented vocabulary used as a qualitative coverage check for scenario terms, functional-zone cues, maritime objects, and surrounding coastal evidence.}
\label{fig:harbor_vocabulary}
\end{figure}

Figure~\ref{fig:harbor_vocabulary} illustrates the semantic breadth of the harbor-oriented data. The curated harbor source collection covers 19 object categories in RGB, 11 in SAR, 12 in PAN, and 9 in NIR. The vocabulary spans vessels and cargo, port infrastructure such as piers, cranes, and storage yards, functional areas such as terminals and anchor zones, and surrounding coastal features such as shorelines, tidal flats, and wetlands. This coverage supports supervision that connects individual objects with spatial relations and functional context. Combined with the staged teacher review described above, it provides diverse, evidence-grounded instruction data for harbor-scenario understanding.

\Needspace{8\baselineskip}
\subsection{Convergent Staged Training}
\label{sec:staged_training}
\begin{figure*}[!t]
\centering
\begingroup
\setlength{\unitlength}{\textwidth}
\begin{picture}(1,0.43208)
\put(0,0){\includegraphics[width=\textwidth,keepaspectratio]{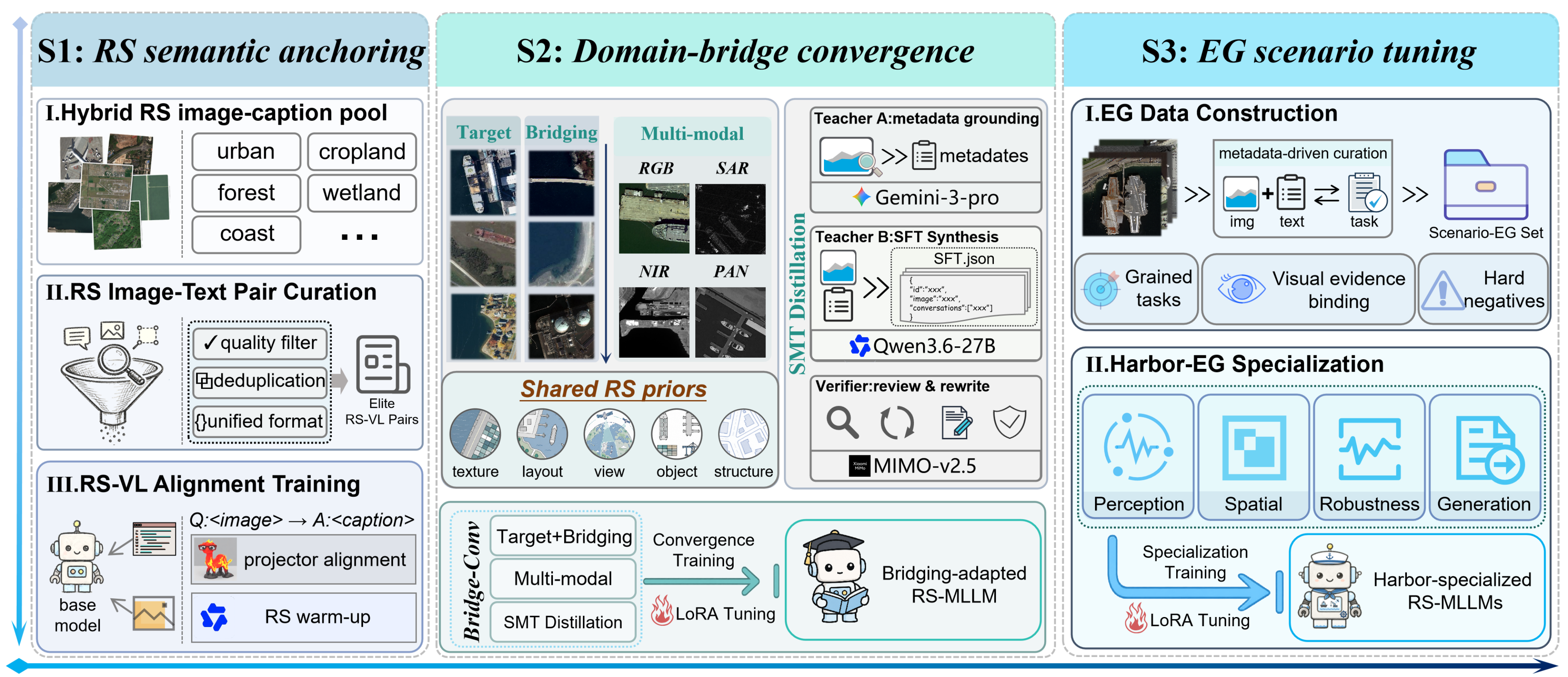}}
\put(0.292,0.021){\rule{0.019\textwidth}{0.081\textwidth}}
\put(0.302,0.062){\makebox(0,0){\rotatebox{90}{\fontsize{7}{7}\selectfont\bfseries\itshape{}Bridging-Conv}}}
\end{picture}
\endgroup
\caption{Convergent staged post-training route: S1 RS semantic anchoring, S2 domain-bridge convergence, and S3 evidence-grounded scenario tuning.}

\label{fig:staged_training_paradigm}
\end{figure*}

Given the three CPRS supervision layers $D_1$, $D_2$, and $D_3$, \textit{FBA} organizes model adaptation into three ordered training stages, denoted by $S_1$, $S_2$, and $S_3$ (Fig.~\ref{fig:staged_training_paradigm}). Let $\mathcal{M}_0$ denote the initial multimodal configuration and $\mathcal{M}_k$ the model obtained after stage $S_k$:
\nopagebreak[4]
\begin{equation}
\label{eq:staged_training}
\mathcal{M}_k=\operatorname{Adapt}(\mathcal{M}_{k-1},D_k),\qquad k\in\{1,2,3\},
\end{equation}
where each stage initializes from the model produced by the preceding stage and uses its corresponding supervision layer. The operator $\operatorname{Adapt}(\cdot,\cdot)$ denotes parameter updates under the training configuration specified for each stage.

At stage $S_1$, RS semantic anchoring adapts $\mathcal{M}_0$ using RS-Anchor supervision $D_1$. This stage establishes the overhead-view visual-language correspondence for RS objects and scene structure, producing $\mathcal{M}_1$ with a broad RS semantic basis for multi-source bridging and scenario tuning. Training lasts one epoch: LLaVA aligns its projector with the vision and language backbones fixed, whereas Qwen3-VL uses language-side LoRA~\citep{hu2022lora} to adapt its instruction-tuned representations.

With the broad RS semantic basis established at $S_1$, stage $S_2$ performs domain-bridge convergence by further adapting $\mathcal{M}_1$ using Bridging-Conv supervision $D_2$. This stage concurrently exposes the model to target harbor scenes and related bridging scenes across multiple modalities. LLaVA updates the projector and LoRA adapters, while Qwen3-VL continues LoRA adaptation for one epoch.

During the $S_2$ update in Eq.~\ref{eq:staged_training}, joint supervision consolidates RS priors shared by target and bridging scenes across modalities, conceptually expressed as
\begin{equation}
\label{eq:bridge_transfer}
\begin{aligned}
&D_{2,\mathrm{t}}^{m}
=
\left\{
(x_i,u_i)\in D_2
\mid
m_i=m,\;
s_i=\mathrm{target}
\right\},\\
&D_{2,\mathrm{b}}^{m}
=
\left\{
(x_i,u_i)\in D_2
\mid
m_i=m,\;
s_i=\mathrm{bridge}
\right\},\\
&\mathcal{P}_{\mathrm{shared}}
=
\operatorname{Shared}_{\mathrm{RS}}
\left(
\left\{
D_{2,\mathrm{t}}^{m},
D_{2,\mathrm{b}}^{m}
\right\}_{m\in\Omega}
\right),
\end{aligned}
\end{equation}
Here, $\Omega=\{\mathrm{RGB},\mathrm{SAR},\mathrm{PAN},\mathrm{NIR}\}$ is the modality set. The subsets $D_{2,\mathrm{t}}^{m}$ and $D_{2,\mathrm{b}}^{m}$ collect verified records from $D_2$ in Eq.~\ref{eq:bridge_curation} with modality label $m_i=m$ and scene labels $s_i=\mathrm{target}$ and $s_i=\mathrm{bridge}$, respectively. The operator $\operatorname{Shared}_{\mathrm{RS}}$ summarizes their common RS priors $\mathcal{P}_{\mathrm{shared}}$, including texture, layout, viewpoint, object patterns, and scene structure.

Bridging-Conv combines target-related context with neighboring coastal and maritime scenes, while broadening the sensor coverage beyond the RGB anchor. For example, the coastal ship imagery in Bridging-Conv shares vessels, shorelines, and spatial relations between vessels and waterfronts with harbor scenes. Learning these objects and relations under different sensor constraints prepares the model to interpret their organization around docks and port facilities during scenario tuning. The resulting $\mathcal{M}_2$ supplies the initialization for scenario tuning.

Finally, stage $S_3$ applies evidence-grounded scenario tuning to $\mathcal{M}_2$ using Scenario-EG supervision $D_3$. This stage focuses training on harbor-specific downstream tasks across perception, spatial understanding, robustness, and generation, while strengthening the use of visual evidence and the handling of hard negative cases. Both backbones use language-side LoRA, while the LLaVA projector is fixed. Stage-specific training settings are provided in Supplementary Table~\ref{tab:training_configuration}. The resulting $\mathcal{M}_3$ is the final harbor-specialized RS-MLLM.

\subsection{Harbor-Scenario Evaluation Protocol}
\label{sec:evaluation_protocol}
We evaluate FBA at two levels. RS-VL Val. and MultiSource Val. diagnose intermediate capabilities acquired along the staged route. Final harbor-scenario performance is assessed using HarborEval, the public-source VRSBench and RSVQA harbor subsets~\citep{li2024vrsbench,lobry2020rsvqa}, and expert-scored OpenEval. Public Harbor in Table~\ref{tab:appendix_evaluation_suite_overview} comprises the two public-source subsets, with results reported separately for each benchmark. Scores are presented on a 0--100 scale to two decimal places. Construction and scoring details are provided in Supplementary Sections~\ref{app:evaluation_suite_construction_audit},
\ref{app:evaluation_protocol_scoring}, and~\ref{app:prompt_rubric_templates}.

\begin{table}[!htbp]

\centering
\caption{Evaluation packages used throughout the paper. Stage diagnostics measure intermediate capability acquisition. HarborEval, Public Harbor, and OpenEval assess final scenario behavior and generalization.}
\label{tab:appendix_evaluation_suite_overview}
\footnotesize
\setlength{\tabcolsep}{1.5pt}
\renewcommand{\arraystretch}{1.12}
\begin{tabularx}{\columnwidth}{@{}>{\raggedright\arraybackslash}p{0.24\columnwidth}>{\raggedright\arraybackslash}p{0.36\columnwidth}>{\raggedright\arraybackslash}X@{}}
\toprule
\rowcolor[HTML]{F2F2F2}
\textbf{Package} & \textbf{Scale} & \textbf{Core purpose} \\
\midrule
RS-VL Val. & 356 images & RS semantic alignment \\
MultiSource Val. & 320 images & Multi-source understanding \\
\midrule
Public Harbor & 580 samples; 200~images & Public-source transfer \\
OpenEval & 100 items & Expert-scored reporting \\
\midrule
\rowcolor[HTML]{E5FFFF}
\textbf{HarborEval} & \textbf{1,245 items; 471~images} & \textbf{Harbor-scenario diagnosis} \\
\bottomrule
\end{tabularx}

\end{table}

\begin{table}[!htbp]

\centering
\caption{HarborEval track composition. ML denotes multi-label, MC multiple-choice, and Y/N/U yes/no/unknown.}
\label{tab:appendix_harboreval_track_distribution}
\footnotesize
\setlength{\tabcolsep}{1.5pt}
\renewcommand{\arraystretch}{1.12}
\begin{tabularx}{\columnwidth}{@{}>{\centering\arraybackslash}p{0.08\columnwidth}>{\raggedright\arraybackslash}p{0.35\columnwidth}>{\centering\arraybackslash}p{0.08\columnwidth}>{\raggedright\arraybackslash}X@{}}
\toprule
\rowcolor[HTML]{F2F2F2}
\textbf{Track} & \textbf{Role} & \textbf{Items} & \textbf{Question types} \\
\midrule
T1 & Object/scene VQA & 162 & 95 ML; 51 MC; 16 Y/N/U \\
\specialrule{0.4pt}{0pt}{0pt}
T2 & Functional zone & 182 & MC \\
\specialrule{0.4pt}{0pt}{0pt}
T3 & Relation reasoning & 183 & 165 MC; 18 Y/N/U \\
\specialrule{0.4pt}{0pt}{0pt}
T4 & Grid grounding & 164 & Grid-choice \\
\specialrule{0.4pt}{0pt}{0pt}
T5 & Observability & 171 & Y/N/U \\
\specialrule{0.4pt}{0pt}{0pt}
T6 & Evidence judgment & 123 & Y/N/U \\
\specialrule{0.4pt}{0pt}{0pt}
T7 & Report generation & 79 & Open-ended \\
\specialrule{0.4pt}{0pt}{0pt}
T8 & Non-harbor rejection & 181 & 113 MC; 56 Y/N/U; 12 open-ended \\
\bottomrule
\end{tabularx}

\end{table}

\begin{table}[!htbp]

\centering
\caption{Open-ended reporting rubric for expert OpenEval scoring and frozen T7 judging. Every dimension is scored on a 0--5 scale. Forbidden claims are flagged separately.}
\label{tab:appendix_openeval_rubric}
\footnotesize
\setlength{\tabcolsep}{1.5pt}
\renewcommand{\arraystretch}{1.12}
\renewcommand{\tabularxcolumn}[1]{m{#1}}
\begin{tabularx}{\columnwidth}{@{}>{\centering\arraybackslash}m{0.25\columnwidth}>{\centering\arraybackslash}m{0.34\columnwidth}>{\centering\arraybackslash}X@{}}
\toprule
\rowcolor[HTML]{F2F2F2}
\textbf{Dimension} & \textbf{Reward} & \textbf{Penalty} \\
\midrule
Groundedness
& Visible or strongly supported evidence
& Unsupported specifics \\
\specialrule{0.4pt}{0pt}{0pt}
Object coverage
& Relevant visible objects
& Missing dominant evidence, invented objects \\
\specialrule{0.4pt}{0pt}{0pt}
Functional-zone understanding
& Evidence-supported zone interpretation
& Overconfident labels for ambiguous regions \\
\specialrule{0.4pt}{0pt}{0pt}
Spatial relation
& Correct layout and object--region relations
& Reversed locations, overstated geometry \\
\specialrule{0.4pt}{0pt}{0pt}
Uncertainty and rejection
& Uncertainty or justified rejection
& Forced harbor claims \\
\specialrule{0.4pt}{0pt}{0pt}
Concision
& Useful, compact report
& Generic verbosity, speculation, repetition \\
\bottomrule
\end{tabularx}

\end{table}

\subsubsection{Intermediate Capability Validation}
\label{sec:intermediate_validation}
RS-VL Val. diagnoses overhead semantic alignment using 356 RGB images. Its primary score equally weights caption-retrieval R@1 and length-matched pairwise accuracy. MultiSource Val. uses 320 images across RGB, SAR, PAN, and NIR to assess sensor-aware descriptions without supplying the modality label. Its fixed image-grounded judge scores six dimensions, aggregates them within each modality, and averages the modality scores. Both sets use fixed inputs and scoring rules across checkpoints. Component definitions, weights, and auxiliary diagnostics are given in Supplementary Section~\ref{app:stage_validation}.

\subsubsection{HarborEval for Scenario Diagnosis}
\label{sec:harboreval}
HarborEval is an eight-track diagnostic benchmark spanning RGB, SAR, PAN, and NIR imagery from harbor and non-harbor scenes, with source records disjoint from CPRS training supervision. It assesses perception, spatial understanding, robustness, and generation through object recognition, functional-zone understanding, modality-dependent observability, spatial relations, grid localization, evidence judgment, non-harbor rejection, and evidence-grounded reporting.

HarborEval contains 1,245 items over 471 images, organized into eight diagnostic tracks (Table~\ref{tab:appendix_harboreval_track_distribution}). Its source records and derived conversations are reserved for evaluation. Each item provides an image, a question, and applicable answer choices.

T1 evaluates object and scene understanding, using set F1 for multi-label questions and accuracy for multiple-choice and yes/no/unknown questions. T2 and T3 assess functional-zone understanding and relation reasoning, respectively, using accuracy against audited reference answers. T4 evaluates image-relative localization using soft grid F1 on a $3\times3$ grid, assigning partial credit to predictions in adjacent cells. The matching procedure and metric definitions are provided in Supplementary Section~\ref{app:evaluation_protocol_scoring} and Eq.~\ref{eq:soft_grid_f1}.

T5 and T6 assess modality-dependent observability and visual claim support, respectively. Both use the responses \textit{Yes}, \textit{No}, and \textit{Cannot determine}, scored by accuracy against audited reference answers. T7 evaluates evidence-grounded report generation using a fixed image-grounded judge. T8 assesses evidence-based rejection of harbor interpretations in non-harbor or near-domain scenes through structured and open-ended questions.

Item-level results are aggregated within each diagnostic track and reported on a common 0--100 scale. Supplementary Section~\ref{app:evaluation_protocol_scoring} provides response normalization, scoring details, and auxiliary diagnostics.

The overall HarborEval score $H_{\mathrm{eval}}$ is computed as the unweighted average of all eight diagnostic tracks after their normalization to a common 0--100 scale:
\begin{equation}
\label{eq:harboreval_score}
H_{\mathrm{eval}}
=
\frac{1}{N_\mathcal{T}}\sum_{t\in\mathcal{T}} h_t,
\quad
\mathcal{T}=\{1,\ldots,N_\mathcal{T}\},
\end{equation}
where $\mathcal{T}$ indexes the diagnostic tracks, $N_{\mathcal{T}}=|\mathcal{T}|=8$, and $h_t\in[0,100]$ is the mean score for track $t$ on the reporting scale. Thus, tracks contribute equally to $H_{\mathrm{eval}}$ regardless of their item counts.

\subsubsection{Open-ended Reporting Criteria}
\label{sec:open_reporting}

HarborEval's T7 track and OpenEval assess evidence-grounded reporting using the six criteria in Table~\ref{tab:appendix_openeval_rubric}, each scored on a 0--5 scale, with forbidden claims recorded during review. T7 uses a fixed image-grounded judge, Doubao-Seed-1.8-Vision~\citep{bytedance2025seed18}, following the model-as-judge approach~\citep{zheng2023judging}. OpenEval uses domain experts, who review responses with model identities and training routes hidden. Borderline or inconsistent cases receive additional review. Supplementary Sections~\ref{app:evaluation_protocol_scoring} and~\ref{app:prompt_rubric_templates} provide the evaluation inputs and detailed scoring criteria.

\section{Experiments and Analysis}

\subsection{Experimental Setup}
\label{sec:experimental_setup}

We evaluate \textit{FBA} on LLaVA-v1.5 and Qwen3-VL~\citep{liu2023visual,liu2024improved,bai2025qwen3vl}, comparing direct and collapsed training with the FBA route described in Section~\ref{sec:staged_training}. The official LLaVA-v1.5 model is also reported as a backbone reference.

The LLaVA-v1.5 experiments use Vicuna-7B, a CLIP ViT-L/14-336 vision tower~\citep{radford2021learning}, and a LLaVA-compatible multimodal projector. The Qwen3-VL experiments use Qwen3-VL-8B with its built-in multimodal processor.

For FBA, $I_0$ denotes the pre-alignment LLaVA configuration, whose projector is aligned during $S_1$, and $B_0$ denotes the released visual-instruction-tuned Qwen3-VL model. The natural-image-aligned LLaVA checkpoint, denoted by $G_0$, initializes Direct-SFT and the collapsed route without $S_1$.

Direct-SFT adapts $G_0$ for LLaVA or $B_0$ for Qwen3-VL directly on $D_3$. Collapsed-SFT combines the bridging and scenario supervision into $C_{2,3}=D_2\cup D_3$ and trains on this pool in a single phase, either from $G_0$/$B_0$ or after $S_1$.

Within each backbone family, Direct-SFT, Collapsed-SFT, and FBA share the same Scenario-EG supervision $D_3$ for final harbor-specialization tasks. FBA adds prerequisite training with $D_1$ and $D_2$ before adaptation on $D_3$. The routes differ in initialization, overall data exposure, and training budget. Detailed configurations are provided in Supplementary Table~\ref{tab:training_configuration} and Section~\ref{app:implementation_reproducibility}.

The stage-specific trainable components follow Section~\ref{sec:staged_training}. For both backbones, $S_1$ and $S_2$ each run for one epoch, while FBA stage $S_3$ and Direct-SFT on $D_3$ run for two epochs. The collapsed routes also use two epochs of $D_3$ exposure, and the anchored collapsed route additionally includes a preceding one-epoch $S_1$ stage. Training uses AdamW in bf16 precision, a default maximum sequence length of 2048 (4096 for LLaVA stage $S_2$), and an effective batch size of approximately 16. Detailed stage-wise learning rates, adapter settings, and batch configurations are reported in Supplementary Table~\ref{tab:training_configuration} and Section~\ref{app:implementation_reproducibility}.

All controlled runs use fixed data manifests and a fixed random seed (default 42). Training exposure is fixed in advance, without validation-based checkpoint selection. Except when resuming an interrupted run, the reported model is the final checkpoint after the planned epochs. Intermediate checkpoints are retained for recovery and audit. Each run uses one NVIDIA RTX A6000 48GB GPU, with two GPUs used only for independent experiments in parallel.

LLaVA-v1.5 and Qwen3-VL are evaluated on HarborEval using the same task prompts and deterministic decoding, with sampling disabled, temperature 0, and beam size 1. External RS-MLLMs use their official checkpoints and recommended prompts when available. All models are evaluated on the same benchmark items, and predictions are normalized to a common scoring format before aggregation.

\label{sec:public_expert}
Public Harbor evaluates captioning and VQA on harbor-related subsets of VRSBench and RSVQA~\citep{li2024vrsbench,lobry2020rsvqa}, comprising 580 samples drawn from 200 images. These subsets assess generalization to additional public-source harbor imagery, with scores reported separately for each benchmark. Supplementary Section~\ref{app:evaluation_protocol_scoring} details the scoring procedures and package-level aggregation.

OpenEval complements these public-benchmark evaluations with 100 expert-assessed items spanning RGB, SAR, PAN, and NIR. It focuses on evidence-grounded reporting and scenario rejection.

\begin{table*}[t]
\centering
\caption{HarborEval diagnostic results for scenario-specialized RS-MLLMs under different training routes.}
\label{tab:harboreval_fine_grained}
\small
\setlength{\tabcolsep}{2.0pt}
\renewcommand{\arraystretch}{1.06}
\begin{tabularx}{\textwidth}{@{}>{\raggedright\arraybackslash}p{0.090\textwidth}>{\raggedright\arraybackslash}p{0.215\textwidth}*{9}{>{\centering\arraybackslash}X}@{}}
\toprule
\rowcolor[HTML]{F2F2F2}
\textbf{Backbone} &
\textbf{Training Route}
& \multicolumn{1}{c}{\textbf{Main}}
& \multicolumn{3}{c}{\textbf{Perception}}
& \multicolumn{2}{c}{\textbf{Spatial}}
& \multicolumn{2}{c}{\textbf{Robustness}}
& \multicolumn{1}{c}{\textbf{Generation}} \\
\cmidrule(lr){3-3}\cmidrule(lr){4-6}\cmidrule(lr){7-8}\cmidrule(lr){9-10}\cmidrule(lr){11-11}
\rowcolor[HTML]{F2F2F2}
&
& Overall
& Object
& Zone
& Observ.
& Relation
& Grid
& Evidence
& Rejection
& Report \\
\midrule

\multicolumn{11}{c}{\textit{LLaVA-v1.5 full-pipeline validation from multimodal initialization}} \\
\midrule
LLaVA-v1.5
& Official Model (ref.)
& 46.22 & 71.42 & 51.67 & 48.54 & 51.12 & 32.13 & 47.15 & 40.24 & 27.47 \\

& Direct-SFT
& \underline{57.95} & \textbf{75.07} & \underline{66.67} & 50.88 & 60.67 & \textbf{48.37} & 52.03 & 37.28 & \underline{72.60} \\

& $G_0\!\rightarrow\!C_{2,3}$
& 55.34 & 42.41 & 52.22 & \underline{61.40} & \underline{68.09} & \underline{45.86} & \underline{52.85} & 58.69 & 61.20 \\

& $I_0\!\rightarrow\!S_1\!\rightarrow\!C_{2,3}$
& 55.74 & \underline{73.85} & 51.11 & 57.89 & 59.55 & 31.74 & 34.96 & \underline{67.46} & 69.34 \\

\cmidrule(lr){2-11}
\rowcolor[HTML]{E5FFFF}
& \textbf{FBA:} $I_0\!\rightarrow\!S_1\!\rightarrow\!S_2\!\rightarrow\!S_3$
& \textbf{70.29} & 73.47 & \textbf{67.22} & \textbf{80.12} & \textbf{69.10} & 43.82 & \textbf{69.11} & \textbf{85.80} & \textbf{73.70} \\

\midrule
\multicolumn{11}{c}{\textit{Native MLLM adaptation from official visual-instruction-tuned base}} \\
\midrule
Qwen3-VL
& $B_0$
& 70.37 & 80.06 & 74.44 & 76.02 & 40.91 & 65.81 & 65.85 & 94.67 & 65.24 \\

& Direct-SFT
& \underline{81.09} & 87.99 & 78.33 & 81.29 & 81.46 & 66.06 & \underline{78.05} & 95.27 & \underline{80.23} \\

& $B_0\!\rightarrow\!C_{2,3}$
& 72.14 & \underline{88.06} & 39.44 & 58.48 & 82.58 & 65.70 & 74.80 & \textbf{98.82} & 69.25 \\

& $B_0\!\rightarrow\!S_1\!\rightarrow\!C_{2,3}$
& 79.36 & 82.06 & \underline{80.56} & \underline{81.97} & \textbf{84.27} & \underline{67.56} & 65.85 & \textbf{98.82} & 73.76 \\

\cmidrule(lr){2-11}
\rowcolor[HTML]{E5FFFF}
& \textbf{FBA:} $B_0\!\rightarrow\!S_1\!\rightarrow\!S_2\!\rightarrow\!S_3$
& \textbf{83.37} & \textbf{92.42} & \textbf{81.11} & \textbf{82.46} & \underline{83.15} & \textbf{68.61} & \textbf{79.67} & \underline{98.22} & \textbf{81.32} \\

\bottomrule
\end{tabularx}

\end{table*}

\subsection{Main Results and Model Comparison}
\subsubsection{Training Route Comparison}

To evaluate scenario specialization under a shared final-task data allocation, we compare FBA with direct and collapsed training routes across the two backbone families. As shown in Table~\ref{tab:harboreval_fine_grained}, FBA raises the overall HarborEval score from 57.95 to 70.29 on LLaVA-v1.5 and from 81.09 to 83.37 on Qwen3-VL relative to Direct-SFT.

For LLaVA-v1.5, observability increases from 50.88 to 80.12, evidence judgment from 52.03 to 69.11, and non-harbor rejection from 37.28 to 85.80. These gains concentrate on evidence-sensitive decisions, including interpreting sensor-constrained observations, assessing visual support for claims, and determining when a harbor interpretation is justified. The pattern suggests a stronger contribution to reliable scenario decisions than to uniform improvement across all fine-grained perceptual abilities.

Qwen3-VL improves across all eight tracks relative to Direct-SFT. Object recognition rises from 87.99 to 92.42, non-harbor rejection from 95.27 to 98.22, and functional-zone understanding from 78.33 to 81.11. Its smaller overall improvement than LLaVA-v1.5 may partly reflect the stronger starting performance and more limited headroom of its instruction-tuned backbone.

On LLaVA-v1.5, FBA reaches an overall score of 70.29, compared with 55.34 and 55.74 for the two collapsed alternatives. On Qwen3-VL, the anchored collapsed route retains $S_1$ before combining the remaining supervision and reaches 79.36, while FBA achieves 83.37.

\begin{table*}[t]
\centering
\caption{Scenario-level comparison with existing RS-MLLMs on HarborEval, VRSBench, RSVQA, and OpenEval.}
\label{tab:rsmllm_compare}
\small
\setlength{\tabcolsep}{1.6mm}
\renewcommand{\arraystretch}{1.05}
\begin{tabularx}{\textwidth}{
>{\raggedright\arraybackslash}p{0.29\textwidth}
>{\centering\arraybackslash}p{0.065\textwidth}
>{\centering\arraybackslash}p{0.13\textwidth}
*{4}{>{\centering\arraybackslash}X}
}
\toprule
\rowcolor[HTML]{F2F2F2}
Model & Params
& Data Scale
& HarborEval
& VRSBench
& RSVQA
& OpenEval \\
\midrule

\multicolumn{7}{c}{\textit{Existing RS-MLLMs}} \\
\midrule
GeoChat\,\textnormal{[CVPR'24]}         & 7B & 318K\textsuperscript{*}              & 47.49 & 53.44 & 51.46 & 21.78 \\
VHM\,\textnormal{[AAAI'25]} & 7B & 1.58M & 57.62 & 49.70 & 52.13 & 54.44 \\
SkyEyeGPT\,\textnormal{[ISPRS'25]}       & 7B & 968K    & 28.28 & 36.72 & 36.72 & 12.33 \\
LHRS-Bot-Nova\,\textnormal{[ISPRS'25]}    & 8B & 2.02M                   & 39.73 & 28.40 & 31.15 & 22.96 \\
SkySenseGPT\,\textnormal{[ISPRS'26]}      & 7B & 3.00M                   & 47.24 & 42.99 & 42.98 & 35.78 \\

\midrule
\multicolumn{7}{c}{\textit{Models trained with the proposed paradigm}} \\
\midrule
\rowcolor[HTML]{E5FFFF}
\textbf{FBA (LLaVA-v1.5)}        & 7B & \textbf{810K}           & \underline{70.29} & \underline{57.62} & \underline{53.96} & \underline{61.47} \\
\rowcolor[HTML]{E5FFFF}
\textbf{FBA (Qwen3-VL)}     & 8B & \textbf{810K}           & \textbf{83.37}    & \textbf{67.77}    & \textbf{63.00} & \textbf{76.67} \\
\bottomrule
\end{tabularx}

\par\vspace{3pt}
\begin{minipage}{\textwidth}
\small 
\textsuperscript{*}GeoChat is initialized from LLaVA-1.5 and its reported data scale excludes the inherited $\sim$1.22M general-purpose samples for consistency.
\end{minipage}

\end{table*}

\subsubsection{Comparison with Existing RS-MLLMs}

We next compare the resulting models with representative RS-MLLMs~\citep{kuckreja2024geochat,zhan2025skyeyegpt,li2025lhrsbotnova,luo2024skysensegpt,pang2025vhm} on HarborEval, the harbor-related subsets of VRSBench and RSVQA, and OpenEval. As shown in Table~\ref{tab:rsmllm_compare}, both \textit{FBA} variants achieve higher scores than the evaluated external RS-MLLMs across all four settings, with FBA-Qwen3-VL attaining the highest score on each benchmark. The FBA models use approximately 810K curated supervision records spanning RS semantic anchoring, multi-source bridging, and evidence-grounded scenario tuning.

\subsection{Stage-wise and Role Analysis}

\begin{table*}[t]
\centering
\caption{Stage-wise capability trajectory. $+S_k$ denotes cumulative training. RS-VL Val. and MultiSource Val. are intermediate diagnostics. HarborEval, VRSBench, RSVQA, and OpenEval assess final harbor-scenario performance.}
\label{tab:stage_trajectory}
\small
\setlength{\tabcolsep}{3pt}
\renewcommand{\arraystretch}{1.08}
\begin{tabularx}{\textwidth}{
@{}
>{\centering\arraybackslash}p{0.17\textwidth}
>{\centering\arraybackslash}p{0.10\textwidth}
>{\hsize=1.12\hsize\linewidth=\hsize\centering\arraybackslash}X
>{\hsize=1.35\hsize\linewidth=\hsize\centering\arraybackslash}X
>{\hsize=0.95\hsize\linewidth=\hsize\centering\arraybackslash}X
>{\hsize=0.92\hsize\linewidth=\hsize\centering\arraybackslash}X
>{\hsize=0.76\hsize\linewidth=\hsize\centering\arraybackslash}X
>{\hsize=0.90\hsize\linewidth=\hsize\centering\arraybackslash}X
@{}
}
\toprule
\rowcolor[HTML]{F2F2F2}
\textbf{Backbone} & \textbf{Route}
& \textbf{RS-VL Val.}
& \textbf{MultiSource Val.}
& \textbf{HarborEval}
& \textbf{VRSBench}
& \textbf{RSVQA}
& \textbf{OpenEval} \\
\midrule

\multirow{4}{*}{\textbf{LLaVA-v1.5}}
& $I_0$
& N/A & N/A & N/A & N/A & N/A & N/A \\

& $+S_1$
& 69.71 & 52.97 & 29.44 & 49.54 & 37.00 & 42.23 \\

& $+S_2$
& \underline{87.53} & \textbf{73.55} & \underline{35.61} & \underline{54.88} & 47.55 & 58.01 \\

& \cellcolor[HTML]{E5FFFF} $+S_3$
& \cellcolor[HTML]{E5FFFF} \textbf{89.16} & \cellcolor[HTML]{E5FFFF} \underline{68.04} & \cellcolor[HTML]{E5FFFF} \textbf{70.29} & \cellcolor[HTML]{E5FFFF} \textbf{57.62} & \cellcolor[HTML]{E5FFFF} \textbf{53.96} & \cellcolor[HTML]{E5FFFF} \textbf{61.47} \\

\midrule

\multirow{4}{*}{\textbf{Qwen3-VL}}
& $B_0$
& 90.40 & 69.60 & 70.37 & 57.02 & 50.63 & 65.48 \\

& $+S_1$
& \textbf{95.29} & 70.32 & \underline{77.26} & 51.14 & 54.66 & 60.03 \\

& $+S_2$
& \underline{93.37} & \textbf{79.77} & 71.80 & \underline{66.58} & 57.30 & 60.15 \\

& \cellcolor[HTML]{E5FFFF} $+S_3$
& \cellcolor[HTML]{E5FFFF} 92.22 & \cellcolor[HTML]{E5FFFF} \underline{76.54} & \cellcolor[HTML]{E5FFFF} \textbf{83.37} & \cellcolor[HTML]{E5FFFF} \textbf{67.77} & \cellcolor[HTML]{E5FFFF} \textbf{63.00} & \cellcolor[HTML]{E5FFFF} \textbf{76.67} \\

\bottomrule
\end{tabularx}

\end{table*}

\begin{table*}[t]
\centering
\caption{LLaVA-side role-replacement controls for testing capability-role assignment. Gen-IT denotes generic image-text data.}
\label{tab:llava_role_replacement}
\small
\setlength{\tabcolsep}{3pt}
\renewcommand{\arraystretch}{1.08}
\renewcommand{\tabularxcolumn}[1]{m{#1}}
\begin{tabularx}{\textwidth}{
@{}
>{\centering\arraybackslash}m{0.075\textwidth}
@{\hspace{2pt}}
>{\raggedright\arraybackslash}m{0.24\textwidth}
>{\hsize=1.12\hsize\linewidth=\hsize\centering\arraybackslash}X
>{\hsize=1.35\hsize\linewidth=\hsize\centering\arraybackslash}X
>{\hsize=0.95\hsize\linewidth=\hsize\centering\arraybackslash}X
>{\hsize=0.92\hsize\linewidth=\hsize\centering\arraybackslash}X
>{\hsize=0.76\hsize\linewidth=\hsize\centering\arraybackslash}X
>{\hsize=0.90\hsize\linewidth=\hsize\centering\arraybackslash}X
@{}
}
\toprule
\rowcolor[HTML]{F2F2F2}
\textbf{Ctrl.} & \textbf{Variant}
& \textbf{RS-VL Val.}
& \textbf{MultiSource Val.}
& \textbf{HarborEval}
& \textbf{VRSBench}
& \textbf{RSVQA}
& \textbf{OpenEval} \\
\midrule

\multirow{2}{*}{$D_1$}
& Gen-IT ($D_1^{\mathrm{gen}}$)
& 76.42 & 64.49 & \underline{60.36} & \underline{56.77} & \underline{51.01} & \underline{55.30} \\

& \cellcolor[HTML]{E5FFFF} RS-Anchor ($D_1^{\mathrm{anc}}$)
& \cellcolor[HTML]{E5FFFF} \textbf{89.16} & \cellcolor[HTML]{E5FFFF} \textbf{68.04} & \cellcolor[HTML]{E5FFFF} \textbf{70.29} & \cellcolor[HTML]{E5FFFF} \textbf{57.62} & \cellcolor[HTML]{E5FFFF} \textbf{53.96} & \cellcolor[HTML]{E5FFFF} \textbf{61.47} \\

\midrule

\multirow{2}{*}{$D_2$}
& Non-bridging ($D_2^{\mathrm{ori}}$)
& 84.36 & 66.90 & 57.11 & 54.77 & 48.13 & 52.06 \\

& \cellcolor[HTML]{E5FFFF} Bridging-Conv ($D_2^{\mathrm{brg}}$)
& \cellcolor[HTML]{E5FFFF} \textbf{89.16} & \cellcolor[HTML]{E5FFFF} \textbf{68.04} & \cellcolor[HTML]{E5FFFF} \textbf{70.29} & \cellcolor[HTML]{E5FFFF} \textbf{57.62} & \cellcolor[HTML]{E5FFFF} \textbf{53.96} & \cellcolor[HTML]{E5FFFF} \textbf{61.47} \\

\midrule

\multirow{2}{*}{$D_3$}
& Non-EG ($D_3^{\mathrm{ori}}$)
& \underline{88.97} & \underline{67.78} & 50.12 & 52.98 & 49.36 & 44.11 \\

& \cellcolor[HTML]{E5FFFF} Scenario-EG ($D_3^{\mathrm{eg}}$)
& \cellcolor[HTML]{E5FFFF} \textbf{89.16} & \cellcolor[HTML]{E5FFFF} \textbf{68.04} & \cellcolor[HTML]{E5FFFF} \textbf{70.29} & \cellcolor[HTML]{E5FFFF} \textbf{57.62} & \cellcolor[HTML]{E5FFFF} \textbf{53.96} & \cellcolor[HTML]{E5FFFF} \textbf{61.47} \\

\bottomrule
\end{tabularx}

\end{table*}

To examine capability development along the staged route, we analyze the cumulative models obtained after $S_1$, $S_2$, and $S_3$. Table~\ref{tab:stage_trajectory} reports their intermediate diagnostic scores and final harbor-scenario performance. The LLaVA route starts from the pre-alignment configuration $I_0$, while the Qwen3-VL route builds on the released instruction-tuned model $B_0$.

For LLaVA, stage $S_2$ strengthens both semantic alignment and multi-source understanding: RS-VL increases from 69.71 to 87.53, and MultiSource rises from 52.97 to 73.55. Scores on all four harbor-scenario evaluations also improve. Building on this checkpoint, stage $S_3$ delivers the largest HarborEval gain, raising the score from 35.61 to 70.29, alongside further improvements in RS-VL, VRSBench, RSVQA, and OpenEval.

The Qwen3-VL trajectory reflects the distinct focus of each training stage. Stage $S_1$ strengthens RS semantic alignment, reaching an RS-VL score of 95.29, while $S_2$ achieves the highest MultiSource score of 79.77. Stage $S_3$ further develops harbor-scenario specialization, increasing HarborEval from 71.80 to 83.37 and OpenEval from 60.15 to 76.67.

For both backbones, MultiSource peaks at $S_2$, whereas the four harbor-scenario evaluations perform best after $S_3$. The distinct stage-wise optima reflect a trade-off between broad multi-source understanding and harbor specialization, making the value of the final stage dependent on target-task requirements.

We further perform role-replacement controls on the LLaVA-v1.5 route to explore the contribution of each supervision layer. Table~\ref{tab:llava_role_replacement} compares RS-Anchor with generic image--text supervision. For the other two layers, Non-bridging and Non-EG use supervision from other publicly available related RS datasets in place of Bridging-Conv and Scenario-EG, respectively.

Replacing generic image--text supervision with RS-Anchor increases the reported RS-VL score from 76.42 to 89.16 and HarborEval from 60.36 to 70.29. In the second group, Bridging-Conv raises MultiSource from 66.90 to 68.04 and HarborEval from 57.11 to 70.29. Its final-scenario difference is therefore larger than its difference on the intermediate multi-source diagnostic.

The third group shows the largest HarborEval separation, from 50.12 with non-EG supervision to 70.29 with Scenario-EG, while RS-VL changes only from 88.97 to 89.16. This pattern is consistent with a stronger effect on scenario-specific behavior than on the broad semantic diagnostic.

\FloatBarrier
\subsection{Bridging Transfer Analysis}

\begin{figure}[!tbp]
    \centering
    \includegraphics[width=\columnwidth]{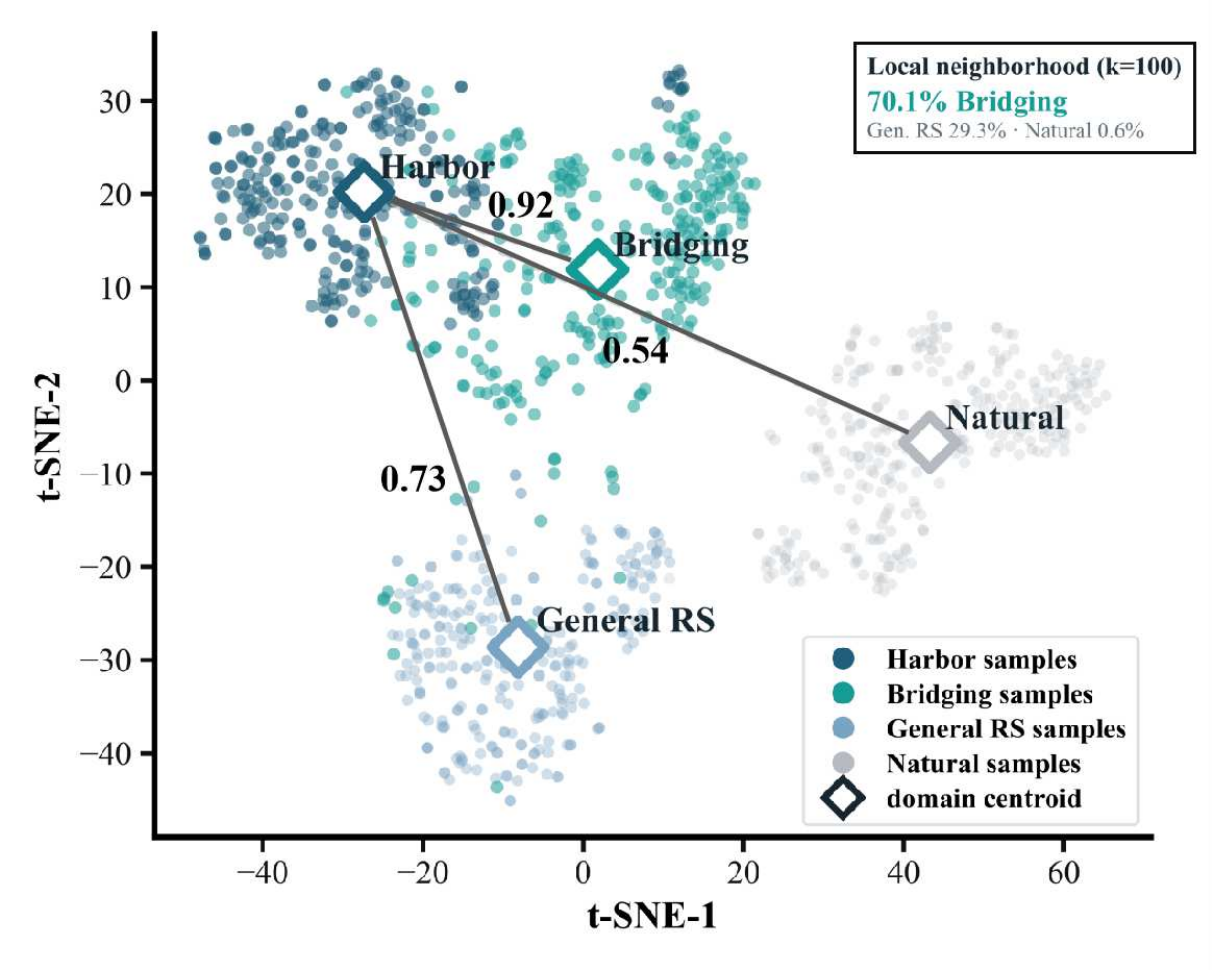}
    \caption{The representation proximity between the Harbor and the other domains. Cosine similarities and kNN statistics are computed in the original joint embedding space.}
    \label{fig:bridging_transfer}
\end{figure}

We assess the suitability of Bridging-Conv scenes as an intermediate supervision domain by measuring their representational proximity to harbor imagery. Harbor, bridging, general RS, and natural-image samples are mapped to a shared visual-language embedding space. Figure~\ref{fig:bridging_transfer} shows their two-dimensional t-SNE distributions~\citep{vanDerMaaten2008tsne}, while centroid cosine similarities and nearest-neighbor statistics are computed in the original embedding space.

In Fig.~\ref{fig:bridging_transfer}, the bridge-domain centroid has a cosine similarity of 0.92 to the harbor centroid, exceeding that of general RS (0.73) and natural-image samples (0.54). Local neighborhoods show a similar pattern. Among the 100 nearest non-harbor neighbors retrieved for each harbor query, bridging samples account for 70.1\%, compared with 29.3\% general RS samples and 0.6\% natural-image samples.

\subsection{Qualitative Evidence-Grounding Analysis}
\label{sec:qualitative_examples}

\begin{figure}[!tb]
\centering
\includegraphics[width=0.96\columnwidth,keepaspectratio]{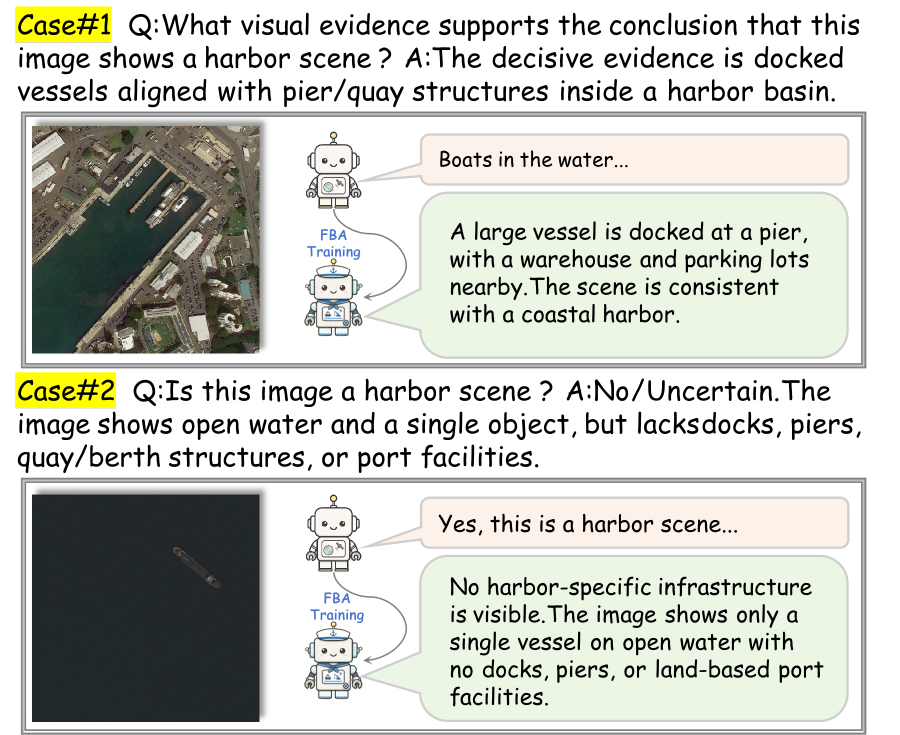}
\caption{Response comparison between a general-purpose MLLM and FBA on harbor and hard-negative cases.}
\label{fig:principle_hardneg}

\end{figure}

Figure~\ref{fig:principle_hardneg} compares responses from a general-purpose MLLM and FBA on a harbor scene and a water-adjacent hard negative. In the harbor case, the general model describes boats in the water, while FBA relates the docked vessel to the pier, warehouse, and surrounding facilities to support the harbor interpretation. In the hard-negative case, FBA identifies an isolated vessel in open water and correctly rejects the general model's harbor prediction.

\begin{figure*}[!t]

\centering
\includegraphics[width=1.0\textwidth,keepaspectratio]{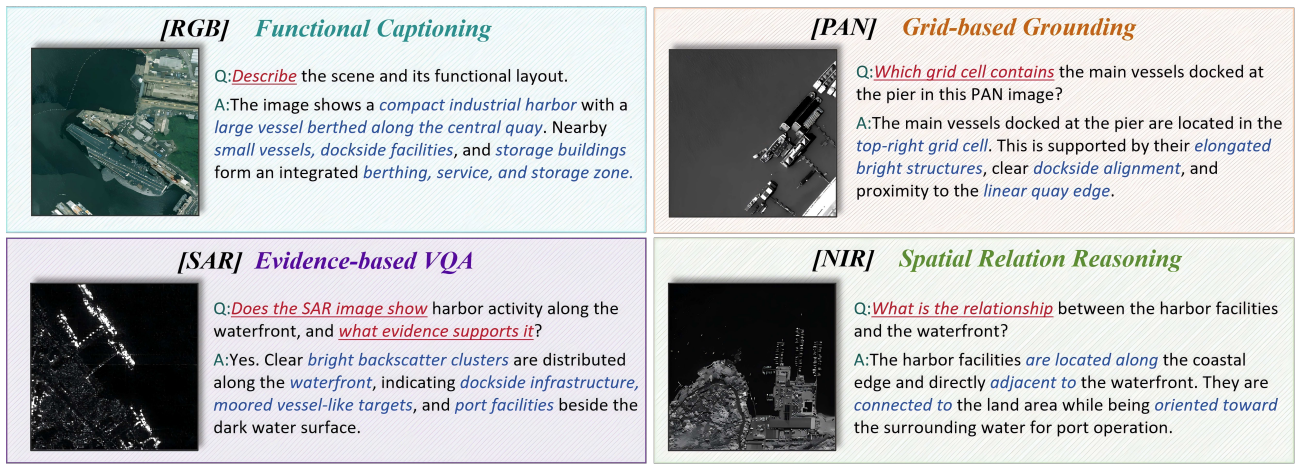}
\caption{Qualitative examples of evidence-grounded harbor understanding across RGB, PAN, SAR, and NIR observations.}
\label{fig:qualitative_harbor_examples}
\end{figure*}

Figure~\ref{fig:qualitative_harbor_examples} illustrates evidence-grounded responses across four modalities. The RGB example links vessels, dockside facilities, and storage buildings to a functional-zone description. Its interpretation depends on how the visible objects are organized around the waterfront, rather than on identifying an isolated ship. The PAN example instead emphasizes grayscale geometry and dockside alignment for image-relative grid localization.

The SAR and NIR examples place greater emphasis on sensor-compatible evidence. The SAR panel uses visible backscatter and spatial structure for an evidence-based answer, while the NIR panel uses waterfront relations to explain the scene. Supplementary Section~\ref{app:qualitative_modality_examples} provides panel-level explanations.

\subsection{Urban-Scenario Extension and Stage-Order Analysis}
\label{sec:urban_extension}

To assess the applicability of FBA beyond harbor scenarios, we conduct a lightweight urban adaptation instantiation using Qwen3-VL-2B-Instruct ($B_0$) as the starting model. We compare different training routes and stage orders to examine how the organization of the three supervision stages affects target-task performance.

\subsubsection{Urban-Scenario Adaptation}
\begin{table*}[!t]
\centering
\caption{Supervision pools and main dataset sources for lightweight urban adaptation.}
\label{tab:urban_supervision_pools}
\small
\setlength{\tabcolsep}{4pt}
\renewcommand{\arraystretch}{1.12}
{
\begin{tabularx}{\textwidth}{@{}>{\raggedright\arraybackslash}m{0.16\textwidth}>{\centering\arraybackslash}m{0.18\textwidth}>{\centering\arraybackslash}m{0.09\textwidth}>{\centering\arraybackslash}m{0.07\textwidth}>{\raggedright\arraybackslash}X@{}}
\toprule
\rowcolor[HTML]{F2F2F2}
\textbf{Pool} & \textbf{Modality} & \textbf{Urban ratio} & \textbf{Scale} & \textbf{Data source} \\
\midrule
RS-Anchor ($D_1$) & RGB & 12.50\% & 80,000 & RSTeller, RSSRData, NWPU-Captions, ChatEarthNet, RSICD, EarthDial, UCM-Captions, Sydney-Captions \\
Bridging-Conv ($D_2$) & RGB / SAR / PAN / NIR & 63.67\% & 10,974 & RSICD, DOTAv2, FAIR1M, DIOR, DIOR-RSVG, SAR-Text, PANBench, SEN12MS, BigEarthNet, EarthDial, EarthGPT \\
Scenario-EG ($D_3$) & RGB & 92.81\% & 6,256 & EarthVQA, EarthVL-GLOBAL, VRSBench-Train, RSVQA-HR \\
\bottomrule
\end{tabularx}
}
\end{table*}

The urban instantiation follows the same supervision-role design, using the data sources and pool configurations summarized in Table~\ref{tab:urban_supervision_pools}. Bridging-Conv reuses urban and related scenes from the existing Stage~2 pool, while Scenario-EG provides task-matched supervision for urban scene understanding.

Evaluation uses subsets of EarthVQA Urban Val, VRSBench-VQA, and RSVQA-HR, with target sizes of 600, 300, and 240 questions, respectively. All evaluation samples are kept separate from the training pools.

As shown in Table~\ref{tab:urban_route_comparison}, FBA achieves the highest scores among the compared training routes across all three evaluations. Relative to $B_0$, the improvement in EarthVQA urban-task performance is accompanied by higher scores on VRSBench-VQA and RSVQA-HR, demonstrating effective urban specialization alongside improvements in other RS question-answering tasks.

\begin{table}[!htbp]
\centering
\caption{Lightweight training-route comparison for urban adaptation.}
\label{tab:urban_route_comparison}
\small
\setlength{\tabcolsep}{2pt}
\renewcommand{\arraystretch}{1.12}
{
\begin{tabularx}{\columnwidth}{@{}
>{\raggedright\arraybackslash}p{0.43\columnwidth}
*{3}{>{\centering\arraybackslash}X}@{}}
\toprule
\textbf{Route}
& \shortstack{\textbf{EarthVQA}\\\textbf{Urban}}
& \shortstack{\textbf{VRSBench}\\\textbf{VQA}}
& \shortstack{\textbf{RSVQA}\\\textbf{HR}} \\
\midrule
$B_0$ & 36.83 & \underline{66.00} & \underline{72.92} \\
Direct-SFT & 63.50 & 63.33 & 69.17 \\
$B_0\rightarrow C_{2,3}$ & 62.50 & 63.67 & 66.67 \\
$B_0\rightarrow S_1\rightarrow C_{2,3}$ & \underline{64.83} & 60.00 & 70.00 \\
\midrule
\rowcolor[HTML]{E5FFFF}
\textbf{FBA:} $\boldsymbol{B_0\rightarrow S_1\rightarrow S_2\rightarrow S_3}$
& \textbf{66.67} & \textbf{67.00} & \textbf{78.33} \\
\bottomrule
\end{tabularx}
}
\end{table}

\subsubsection{Effect of Stage Order}

As shown in Table~\ref{tab:urban_stage_order}, the $S_1\rightarrow S_2\rightarrow S_3$ sequence achieves the strongest overall performance among the compared orders. The $S_2\rightarrow S_1\rightarrow S_3$ comparison examines the order of semantic anchoring and multi-source bridging while keeping scenario tuning last, whereas $S_1\rightarrow S_3\rightarrow S_2$ examines the effect of placing bridging after scenario tuning. These comparisons favor establishing prerequisite capabilities through anchoring and bridging before final scenario specialization.

\begin{table}[!htbp]
\centering
\caption{Stage-order comparison for urban adaptation.}
\label{tab:urban_stage_order}
\small
\setlength{\tabcolsep}{2pt}
\renewcommand{\arraystretch}{1.12}
{
\begin{tabularx}{\columnwidth}{@{}>{\raggedright\arraybackslash}p{0.43\columnwidth}*{3}{>{\centering\arraybackslash}X}@{}}
\toprule
\textbf{Sequence} & \shortstack{\textbf{EarthVQA}\\\textbf{Urban}} & \shortstack{\textbf{VRSBench}\\\textbf{VQA}} & \shortstack{\textbf{RSVQA}\\\textbf{HR}} \\
\midrule
$S_2\rightarrow S_1\rightarrow S_3$ & \underline{61.67} & 51.67 & 48.33 \\
$S_1\rightarrow S_3\rightarrow S_2$ & 60.83 & \underline{62.67} & \underline{73.75} \\
\midrule
\rowcolor[HTML]{E5FFFF}
\textbf{FBA:} $\boldsymbol{S_1\rightarrow S_2\rightarrow S_3}$ & \textbf{66.67} & \textbf{67.00} & \textbf{78.33} \\
\bottomrule
\end{tabularx}
}
\end{table}

\FloatBarrier

\section{Discussion}
\label{sec:discussion}

\subsection{Supervision Design for Scenario Specialization}
FBA uses broader RS supervision to build prerequisite capabilities, letting the final stage focus on scenario-specific decisions and evidence-grounded responses. Bridging-Conv connects overhead semantic understanding with target-scene interpretation through shared objects, spatial relations, and sensor constraints. This organization allows scarce task-matched supervision to build on capabilities developed in the preceding stages.

\subsection{Training Cost and Scope of Evidence}
The additional stages require more training and data-preparation effort than direct tuning on $D_3$. Route selection should therefore reflect available resources and balance broad, multi-source understanding with target-scenario performance. Further investigation can isolate the contribution of individual data-curation choices and assess the stability of the training route across repeated runs.

\subsection{Applicability and Future Directions}
The harbor instantiation provides a detailed FBA instantiation on two backbones, complemented by the lightweight urban instantiation in Section~\ref{sec:urban_extension}. Scenario-specialized supervision and evaluation benchmarks require task-matched imagery, domain expertise, and detailed evidence annotation, making them costly to construct. CPRS and HarborEval address these needs for coastal harbors, covering diverse objects, functional zones, spatial relations, and sensor-dependent evidence. Extending FBA to other applications requires selecting semantic anchors and bridging scenes that share relevant objects, spatial relations, and sensor characteristics with the target tasks, together with task-matched supervision and evaluation.

\section{Conclusion}

We present FBA, a capability-gap-driven post-training route for adapting scenario-specialized RS-MLLMs. FBA develops prerequisite capabilities through RS semantic anchoring and multi-source bridging before evidence-grounded scenario tuning. We instantiate this route for coastal harbor understanding through CPRS and introduce HarborEval, an eight-track diagnostic benchmark covering perception, spatial understanding, robustness, and report generation across RGB, SAR, PAN, and NIR observations.

On LLaVA-v1.5 and Qwen3-VL, FBA achieves higher overall HarborEval scores than direct and collapsed training under the same final-task supervision allocation. Evaluations on public harbor subsets and expert-scored OpenEval further demonstrate its effectiveness in question answering, captioning, and reporting. The lightweight urban extension further shows the paradigm's applicability beyond harbors, with stage-order comparisons favoring semantic anchoring and multi-source bridging before final scenario specialization.

\section*{Declaration of competing interest}

The authors declare that they have no known competing financial interests or personal relationships that could have appeared to influence the work reported in this paper.

\section*{Acknowledgments}

This work is supported by the National Natural Science Foundation of China under grants 62571271 and 62201552, the Natural Science Foundation of Tianjin under Grant No. 24JC\-QNJC01890, and the Fundamental Research Funds for the Central Universities.

\section*{CRediT authorship contribution statement}
\textbf{Yuheng Zong:} Conceptualization, Methodology, Investigation, Writing -- original draft. \textbf{Minghua Wang:} Supervision, Resources, Funding acquisition, Writing -- review \& editing. \textbf{Xin Zhao:} Supervision, Funding acquisition. \textbf{Zhi-Hui Zhan:} Conceptualization, Writing -- review \& editing. \textbf{Antonio Plaza:} Writing -- review \& editing. \textbf{J{\'o}n Atli Benediktsson:} Writing -- review \& editing.

\section*{Data availability}
Public examples and release announcements are available on the
\href{https://z0ngl1ng.github.io/filling-before-advancing/}{%
\textbf{\textit{project website}}}.
Model weights and CPRS data will be released progressively through
\href{https://huggingface.co/zongling/FBA}{%
\textbf{\textit{Hugging Face (models)}}}
and
\href{https://huggingface.co/datasets/zongling/CPRS}{%
\textbf{\textit{Hugging Face (data)}}},
respectively. Third-party data remain subject to their providers' terms.

\FloatBarrier
\label{main:body_end}

\begingroup
\small
\renewcommand{\bibfont}{\fontsize{8}{8.7}\selectfont}
\interlinepenalty=10000
\bibliographystyle{elsarticle-harv}
\bibliography{references}
\endgroup

\flushcolsend
\clearpage
\raggedcolsend
\input{supplement_body}
\FloatBarrier
\end{document}

%% file: preamble.tex
\usepackage[T1]{fontenc}
\usepackage{helvet}
\usepackage{courier}
\usepackage[hyphens]{url}
\usepackage{xurl}
\def\UrlFont{\rm}
\usepackage{graphicx}
\usepackage{caption}
\usepackage{microtype}
\usepackage[nospread]{cuted}
\AtBeginDocument{%
  \lefthyphenmin=2\relax
  \righthyphenmin=3\relax
  \clubpenalty=10000
  \widowpenalty=10000
  \displaywidowpenalty=10000
}
\usepackage{booktabs}
\usepackage{tabularx}
\usepackage{array}
\usepackage{colortbl}
\usepackage{xcolor}
\AtBeginDocument{\arrayrulecolor{black}}
\usepackage{amsmath}
\usepackage{amssymb}
\usepackage{multirow}
\usepackage{comment}
\usepackage{placeins}
\usepackage{needspace}
\usepackage[nospread,noshrink]{flushend}
\usepackage[hidelinks]{hyperref}
\hypersetup{
 pdftitle={Filling Before Advancing: Capability-Gap-Driven Post-Training for Scenario-Specialized Remote Sensing MLLMs},
 pdfauthor={Yuheng Zong; Minghua Wang; Xin Zhao; Zhi-Hui Zhan; Antonio Plaza; Jon Atli Benediktsson},
 pdfsubject={Scenario-Specialized Remote Sensing Multimodal Large Language Models},
 pdfkeywords={remote sensing; multimodal large language models; post-training; scenario specialization}
}
\journal{ISPRS Journal of Photogrammetry and Remote Sensing}
\makeatletter
\let\ps@pprintTitle\ps@plain
\makeatother

%% file: supplement_body.tex
\begin{strip}
\par\vspace{16pt}
\centering
{\LARGE\bfseries Supplementary Material\par}
\vspace{0.6em}
{\large
Filling Before Advancing: Capability-Gap-Driven Post-Training for\par
Scenario-Specialized Remote Sensing MLLMs\par
}
\vspace{0.8em}
\end{strip}

\setcounter{secnumdepth}{2}
\setcounter{section}{0}
\setcounter{subsection}{0}
\setcounter{table}{0}
\setcounter{figure}{0}
\setcounter{equation}{0}

\renewcommand{\thesection}{S\arabic{section}}
\renewcommand{\thesubsection}{S\arabic{section}.\arabic{subsection}}
\renewcommand{\thetable}{S\arabic{table}}
\renewcommand{\thefigure}{S\arabic{figure}}
\renewcommand{\theequation}{S\arabic{equation}}

\providecommand{\theHsection}{}
\providecommand{\theHsubsection}{}
\providecommand{\theHtable}{}
\providecommand{\theHfigure}{}
\providecommand{\theHequation}{}
\renewcommand{\theHsection}{supp.section.\arabic{section}}
\renewcommand{\theHsubsection}{supp.subsection.\arabic{section}.\arabic{subsection}}
\renewcommand{\theHtable}{supp.table.\arabic{table}}
\renewcommand{\theHfigure}{supp.figure.\arabic{figure}}
\renewcommand{\theHequation}{supp.equation.\arabic{equation}}

\section{Additional Data-Curation Details}
\label{app:data_curation}
\subsection{Unified Data Collection and Curation Pipeline}
Although the stages differ in supervision format, they share a common curation pipeline. We normalize heterogeneous sources into image-text or ShareGPT-style instruction records, with explicit image references, modality labels, and task fields when available. For image-text data, we enforce image-level uniqueness and retain one caption per image. For instruction data, we center conversations on visible evidence and exclude audit-only metadata from the final training input.

The main filtering criterion is visual groundedness. We reduce weakly visual expressions, including place names, addresses, precise distances, geographic coordinates, unsupported attributes, and metadata-dependent answers. For non-RGB modalities, we set language constraints in view of sensor-dependent observability~\citep{zhu2021deep,vivone2023multispectral}: synthetic aperture radar (SAR) samples allow conservative uncertainty, near-infrared (NIR) samples avoid RGB color assumptions, and panchromatic (PAN) samples emphasize grayscale contrast, geometry, texture, and spatial organization. Model-assisted rewriting produces descriptions, visual question answering (VQA), localization, region understanding, relation reasoning, modality-aware questions, and concise reports. Fixed teacher roles are used for metadata grounding, instruction synthesis, and evidence verification. Their prompts, generation settings, and retry outcomes are retained only as audit information and are excluded from the exported student records. Subsequent checks remove malformed conversations, duplicated answers, missing image tokens, coordinate-style leakage, and modality-incompatible claims.

\subsection{Stage-Specific Progressive Design}
Stage~1 constructs RS-Anchor, a broad RGB remote-sensing (RS) visual-language anchor. After caption cleaning, image deduplication, scene classification, and diversity-aware sampling, it retains 569,853 unique image-caption pairs from 3,135,250 original samples across eight public datasets~\citep{lu2018exploring,yuan2022exploring,cheng2022nwpu,ge2025rsteller,yuan2025chatearthnet,soni2025earthdial}. The retained captions emphasize observable land cover, land use, spatial layout, object distribution, and coarse scene structure. Metadata-heavy descriptions are filtered to provide visually grounded semantics before instruction tuning and non-RGB exposure. The retained Stage~1 inventory comprises RSTeller (407,566 pairs), RSSRData (95,776), NWPU Caption (29,609), ChatEarthNet subsets (15,278), RSICD (10,219), EarthDial (8,806), UCM Captions (2,039), and Sydney Captions (560). Smaller caption datasets supply targeted coverage where available, while RSTeller preserves scale after strict visual-grounding filters.

Stage~2 constructs Bridging-Conv for domain-bridge convergence. The final curated mixture contains 187,296 supervised fine-tuning (SFT) samples across RGB, SAR, NIR, and PAN. Its RGB subset contains 99,088 image-unique samples, including 59,453 water-, coast-, port-, dock-, or ship-related records. The non-RGB subsets contain 29,984 SAR, 28,475 NIR, and 29,749 PAN SFT records. This bridging stage is scenario-aware without collapsing into the final harbor task. Its underlying sources include SAR text-anchored data, object and ship recognition datasets, multispectral land-cover corpora, public caption and VQA resources, near-domain maritime samples, and additional public harbor imagery. Representative documented sources in these families include OpenSARShip, SEN12MS, BigEarthNet, DOTA, DIOR, and EarthDial~\citep{huang2018opensarship,schmitt2019sen12ms,sumbul2019bigearthnet,xia2018dota,li2020dior,soni2025earthdial}. These sources are not concatenated as raw records; they are converted into image-centered SFT conversations, rewritten under modality constraints, and audited for visual groundedness. The non-RGB subsets follow modality-specific constraints: SAR emphasizes structural layout and uncertainty control, NIR avoids RGB color assumptions, and PAN stresses geometry, edges, grayscale contrast, and spatial organization.

Stage~3 constructs Scenario-EG for evidence-grounded scenario tuning. It contains 53,000 train-only ShareGPT samples associated with 8,703 RGB, SAR, PAN, and NIR images. Its supervision emphasizes relation reasoning, presence validation, multi-cell grid localization, and functional-zone descriptions that distinguish dominant, secondary, mixed, and uncertain interpretations. Controlled non-harbor negatives and short-answer VQA support evidence-based rejection and concise, direct responses. The final pool is selected from a larger intermediate set and audited to remove malformed conversations, missing image references, metadata and task-field leakage, benchmark-specific traces, underscore labels, and unsupported identity, activity, cargo, location, or temporal claims. Main-paper Table~\ref{tab:appendix_data_source_statistics} summarizes the three training-supervision pools; the separately maintained evaluation packages are documented in Section~\ref{app:evaluation_suite_construction_audit}.

\paragraph{Source-family inventory}
The Stage~2 and Stage~3 pools draw on complementary sensor and task families rather than raw benchmark merges. SAR supervision combines text-anchored radar records with OpenSARShip and HRSID for ship, water, port, and structural-layout evidence under conservative uncertainty~\citep{huang2018opensarship,wei2020hrsid}. PAN and multispectral supervision uses PANBench, SEN12MS, and BigEarthNet to emphasize grayscale structure, reflectance, texture, geometry, and land-cover context without importing RGB-only color assumptions~\citep{wang2023panbench,schmitt2019sen12ms,sumbul2019bigearthnet}.

Object- and region-oriented records from DOTA, DIOR, and optical or SAR ship-recognition sources, including ShipRSImageNet, are converted into natural-language localization, relation, and region-understanding conversations~\citep{xia2018dota,li2020dior,zhang2021shiprsimagenet}. EarthDial, EarthGPT, and established RS caption corpora provide description and question-answering material, while MME-RealWorld contributes only evidence-compatible rejection and contrastive repair cases~\citep{soni2025earthdial,zhang2024earthgpt,lu2018exploring,zhang2025mmerealworld}. Additional public harbor and maritime imagery expands variation in dock, quay, coastal industrial, and ship-dense layouts. All retained records undergo the same image deduplication, modality-aware rewriting, SMT verification, leakage removal, and train-only export described above.

\paragraph{Count traceability and data separation}
The reported counts are derived from frozen construction inventories and independently checked against the final exported training pools. For each stage, the audit records preserve source-level counts, image-level deduplication outcomes, modality composition, missing-image checks, and the removal of metadata or benchmark-related fields. Superseded intermediate pools are excluded from the totals, preventing duplicate counting when a sample is rewritten or replaced. Before training, all retained records pass image deduplication, modality-aware rewriting, SMT verification, leakage removal, and train-only export. The resulting pools are therefore curated instruction data rather than raw benchmark merges, while source identifiers and collection notes remain available for internal audit.

\paragraph{Harbor imagery coverage}
The harbor-oriented records used for Bridging-Conv, Scenario-EG, HarborEval, and OpenEval are designed to cover more than a single local port style. As summarized visually in main-paper Fig.~\ref{fig:data_curation_pipeline}(b), the collection spans Pacific and Atlantic coasts, the Gulf of Mexico, Atlantic Europe, the Baltic Sea, the Mediterranean Coast, the East Asia Coast, the South China Sea, and maritime Southeast Asia. Within East Asia, additional representative large-port imagery strengthens examples of coastal industrial harbor layouts, berthing zones, storage yards, inland-water connections, and ship-dense operational scenes. These geographic and functional additions are used as curation diversity rather than as location labels exposed to the student model.

The harbor-oriented vocabulary is summarized in main-paper Fig.~\ref{fig:harbor_vocabulary}, which complements the geographic coverage described above with a qualitative overview of maritime objects, port infrastructure, functional zones, and coastal context.

\Needspace{6\baselineskip}
\section{Bridging-Conv Construction Workflow}
\label{app:smt_workflow}
The main paper states the supervision role of Bridging-Conv; here we give its construction procedure and formal definition. For each source image, the pipeline first identifies the sensor modality and normalizes the available source and scene evidence under modality-specific observability constraints. RGB, SAR, PAN, and NIR inputs are then converted into their corresponding training representations and paired with target or neighboring bridging context, including harbor and port scenes, water, coasts, ships, docks, and related infrastructure. An instruction-synthesis teacher converts this grounded evidence into a candidate image--instruction--answer record, after which a stricter verifier either retains the record, rewrites unsupported wording, or discards the sample. The resulting audited records form the Stage~2 Bridging-Conv supervision pool. Source metadata, construction tags, and verifier rationales support curation and auditing only; they are never exposed to the student model as privileged training inputs. Teacher and verifier identities, prompts, rubrics, and retry rules are retained as implementation records for reproducibility.

The three teacher roles and the retained supervision pool are defined in main-paper Eq.~\ref{eq:bridge_curation}. The implementation details below expand that construction procedure.

\subsection{Staged Multi-Teacher Distillation}
We use staged multi-teacher (SMT) distillation as a construction-time mechanism, not as a separate model component. Its three ordered roles are metadata grounding, instruction synthesis, and evidence verification. Metadata grounding normalizes source and modality tags under sensor-specific observability constraints. Instruction synthesis then produces ShareGPT-style image--instruction--answer records, and the verifier retains, rewrites, or drops candidates under stricter visual-evidence rules. The student sees only the final audited records, while private metadata, verifier rationales, and teacher notes are removed.

For Bridging-Conv, SMT controls hallucination risks introduced by heterogeneous sources and sensor-dependent observability~\citep{li2023evaluating}. For SAR, PAN, and NIR, verification suppresses unsupported color, material, fine-identity, and activity claims. Across modalities, it removes weakly visual answers, metadata-dependent statements, duplicate image references, benchmark traces, and unsupported operation descriptions. Retained samples emphasize observable scene structure, objects, spatial relations, modality-aware evidence, and evidence-compatible uncertainty; unreliable samples are discarded rather than template-expanded.

\begin{table*}[t]
\centering
\caption{Semi-structured SMT prompt sketches for Bridging-Conv construction. The three roles decompose Stage~2 synthesis into evidence grounding, instruction generation, and verifier-based repair. Braced fields are populated during construction; private source fields, verifier rationales, and teacher notes are removed before ShareGPT export.}
\label{tab:appendix_smt_prompt_templates}
\begin{tabularx}{\linewidth}{@{}>{\raggedright\arraybackslash}X@{}}
\toprule
\rowcolor[HTML]{F2F2F2}
{\small\textbf{SMT-P1 Evidence Grounding}}\tabularnewline
{\small\ttfamily Role: evidence-grounding teacher; produce audit tags, not final SFT answers.\newline Inputs: \{image/rendering\}, \{modality: RGB|SAR|PAN|NIR\}, \{source fields\}, \{scenario context\}.\newline Check: visible objects; functional zones; spatial layout; sensor-specific observability; uncertainty risks.\newline Forbid: unsupported location, operator, cargo, identity, time, exact activity, and RGB-only color claims for non-RGB images.\newline Output: \{scene tags, visible evidence, spatial layout, modality limits, forbidden claims, risk flags\}.}\tabularnewline
\midrule
\rowcolor[HTML]{F2F2F2}
{\small\textbf{SMT-P2 Instruction Synthesis}}\tabularnewline
{\small\ttfamily Role: synthesize one ShareGPT-style Bridging-Conv candidate.\newline Inputs: \{image/rendering\}, \{evidence tags\}, \{task role\}, \{modality\}, \{target or neighboring bridging context\}.\newline Task menu: scene/object recognition; relation reasoning; spatial description; modality judgment; evidence-compatible uncertainty.\newline Constraint: ask and answer only from visible, modality-compatible evidence; never expose metadata, labels, coordinates, rubrics, or source fields.\newline Output: \{user turn, assistant turn, task type, modality, evidence summary, construction tags\}.}\tabularnewline
\midrule
\rowcolor[HTML]{F2F2F2}
{\small\textbf{SMT-P3 Verify, Repair, or Drop}}\tabularnewline
{\small\ttfamily Role: strict verifier for evidence-grounded training records.\newline Inputs: \{image/rendering\}, \{candidate SFT record\}, \{evidence tags\}, \{forbidden claims\}.\newline Audit: visual support; modality compatibility; answer length; duplicate image references; benchmark traces; metadata leakage; unsupported identity/cargo/operator/location/time claims.\newline Decision: KEEP if supported; REWRITE if the task is useful but wording is unsafe; DROP if evidence is insufficient or the sample is unrecoverable.\newline Output: \{decision, revised user, revised assistant, evidence status, flags, drop reason\}.}\tabularnewline
\bottomrule
\end{tabularx}

\end{table*}

\section{Evaluation Suite Construction and Audit}
\label{app:evaluation_suite_construction_audit}
\label{app:harboreval_construction_audit}

\subsection{Evaluation Package Roles and Separation}
The harbor study uses five evaluation sets with distinct roles (main-paper Table~\ref{tab:appendix_evaluation_suite_overview}). RS-VL Val. and MultiSource Val. are fixed diagnostics of the capabilities targeted by Stages~1 and~2. HarborEval assesses final harbor-scenario understanding. Public Harbor evaluates transfer to harbor subsets of VRSBench and RSVQA, and OpenEval assesses open-ended, evidence-grounded reporting through expert review. We report the stage diagnostics separately from the final scenario evaluations.

\subsection{HarborEval Construction and Audit}
HarborEval evaluates whether a model specialized for coastal harbor RS can combine object recognition, functional interpretation, spatial reasoning, sensor-aware observability, uncertainty handling, non-harbor rejection, and grounded reporting under the same input restrictions as the other models. We maintain the benchmark separately from all training pools. Scenario-EG is used only for training, whereas HarborEval answers, evidence, accepted labels, and scoring rubrics are retained only for evaluation and auditing. Once a source record is assigned to HarborEval, all conversations derived from it are excluded from training. This separation of source records complements the controls on information exposed during inference.

\subsection{Evaluation Scope and Public/Private Separation}
HarborEval contains 1,245 items over 471 unique images. Among them, 1,154 items are structured closed-form items and 91 items are open-ended description or rejection items. The average number of items per image is 2.64 and the maximum is 7. This item--image structure lets HarborEval probe several capability roles on the same visual source when appropriate, while keeping public inference records separate from answer-bearing audit records.

The benchmark is organized into public inference records and private answer records. During inference, a model receives only the image, question, and answer choices when the task is closed-form. Fields such as modality, scene, difficulty, metric type, track name, answer, reference answer, evidence objects, evidence relations, accepted grid cells, and scoring rubrics are not inserted into the prompt. The track field is retained only for evaluator-side grouping. The field-level audit found no disallowed private fields in the public inference records. This check complements, but does not replace, the source-record holdout described above.

\subsection{Track Construction}
HarborEval contains eight diagnostic tracks. The first seven tracks focus on harbor-scenario understanding and grounded reporting, while the eighth track tests whether the model avoids forcing harbor interpretations onto non-harbor or near-domain scenes. The headline score in the main paper macro-averages all eight tracks after normalizing track-specific metrics to a common scale; for diagnostic interpretation, the T8 rejection track and the T7 open-ended description track are also inspected separately.

The tracks are tied to capability roles rather than to isolated task formats. T1 and T2 measure whether the model recognizes scenario-relevant objects and functional areas. T3 and T4 measure whether it can reason about relative layout and localize evidence without bounding-box supervision. T5 checks whether the model respects sensor-dependent observability constraints across RGB, SAR, PAN, and NIR. T6 tests whether the model can distinguish visible evidence from unsupported claims, including cases where identity, cargo type, operator, exact status, or location cannot be determined from imagery alone. T7 tests whether open-ended reports remain grounded and concise. T8 tests whether the model rejects false harbor interpretations in non-harbor or near-domain scenes.

\subsection{Construction and Cleaning Workflow}
The construction process starts from harbor and non-harbor RS records and converts them into item-level diagnostic questions. For harbor records, we generate or audit object/scene questions, functional-zone labels, relation questions, grid-localization targets, modality-observability questions, evidence-calibrated yes/no/unknown judgments, and open-ended report prompts. For non-harbor records, we construct rejection-oriented questions that require the model to avoid hallucinating docks, quays, cargo terminals, vessels, or port logistics when these elements are not supported by the image.

The cleaning workflow removes weakly visual questions, metadata-dependent answers, benchmark-specific traces, duplicated image references, malformed option sets, and modality-incompatible statements. For non-RGB items, we audit whether the wording relies on visible structure, grayscale contrast, backscatter/texture, reflectance, edges, or layout rather than RGB-only color assumptions. For evidence-calibrated questions, we remove or rewrite prompts that can be answered only from metadata, geographic knowledge, source filenames, hidden labels, or external facts. For open-ended items, forbidden-claim fields are kept in the private answer package to penalize unsupported port names, vessel identities, cargo types, coordinates, operational claims, and temporal claims during scoring.

For visually ambiguous items, we retain alternative answers only when supported by the image evidence. Functional-zone items can accept secondary or mixed labels when more than one coarse interpretation is supported. For relation items, alternatives such as near versus adjacent or docked versus moored are accepted only when visually defensible. Grid-grounding items use sets of accepted cells to accommodate large objects and objects spanning multiple cells.

\subsection{Answer Distribution and Boundary Cases}
The package is designed as a diagnostic benchmark, so answer balance is inspected at the track level rather than assumed globally. Several subtracks have asymmetric labels for substantive reasons: some auxiliary yes/no subsets are biased toward visible evidence, while the rejection track naturally contains more negative decisions because unsupported harbor-specific claims should be refused. We therefore report the headline track macro together with per-track scores and auxiliary strict or open-ended diagnostics instead of relying on a single item-level accuracy number.

\begin{table}[t]
\centering
\caption{Selected HarborEval audit statistics used to interpret diagnostic scores.}
\label{tab:appendix_harboreval_audit_stats}
\small
\setlength{\tabcolsep}{2.5pt}
\renewcommand{\arraystretch}{1.10}
\begin{tabularx}{\columnwidth}{>{\raggedright\arraybackslash}p{0.25\columnwidth}>{\centering\arraybackslash}p{0.16\columnwidth}>{\raggedright\arraybackslash}X}
\toprule
\rowcolor[HTML]{F2F2F2}
\textbf{Audit item} & \textbf{Value} & \textbf{Interpretation} \\
\midrule
Closed and open split & 1,154/91 & Package scope. \\
\specialrule{0.4pt}{0pt}{0pt}
Images and mean items & 471/2.64 & Item scores are correlated by image. \\
\specialrule{0.4pt}{0pt}{0pt}
T5--T6 revised/replaced & 294/64 & Weakly visual items were revised. \\
\specialrule{0.4pt}{0pt}{0pt}
T2 majority-class rate & 32.97\% & Context for functional-zone accuracy. \\
\specialrule{0.4pt}{0pt}{0pt}
T4 multi-correct ratio & 40.24\% & Motivates accepted sets and soft grid F1. \\
\specialrule{0.4pt}{0pt}{0pt}
T7 complete audit packets & 79 & All open reports retain private scoring support. \\
\specialrule{0.4pt}{0pt}{0pt}
Public-field violations & 0 & Private answer and evidence fields are excluded. \\
\bottomrule
\end{tabularx}

\end{table}

T2 contains six coarse functional-zone classes. The largest class is water or navigation area with 60 items, followed by mixed or uncertain port area with 39 items, berthing area with 28 items, cargo storage area with 25 items, marina or small-boat harbor with 16 items, and industrial or logistics area with 14 items. This distribution reflects the visual structure of harbor scenes: water/navigation areas are frequent, but the task still requires distinguishing storage, berthing, industrial/logistics, marina, and mixed-use zones. T4 contains 164 grid-grounding items. Among them, 98 items have a single accepted grid cell and 66 have multiple accepted cells, with an average accepted-cell set size of 1.79 and a maximum of 9. The multi-correct ratio of 40.24\% reflects the fact that vessels, docks, basins, and storage areas often span grid boundaries.

\subsection{Open-ended Audit}
The T7 caption/report track contains 79 open-ended items. Each item includes private scoring support: a scoring rubric, forbidden claims, a reference answer, and key points. The rubric has priority over optional key points during frozen T7 judging. Environmental descriptors such as clear weather, calm sea state, daytime, nighttime, or modality names are treated as optional consistency cues unless explicitly required by the rubric. A model is therefore not penalized solely for omitting such cues when it correctly reports visible objects, relations, and layout. Conversely, hallucinated objects, unsupported port names, vessel identities, cargo types, coordinates, or operational details are penalized.

This design is important for RS reporting. Overhead imagery often supports functional and spatial interpretation, but it rarely supports exact vessel identity, cargo type, operator, port name, throughput, or timestamp-level claims. The open-ended audit therefore rewards grounded coverage and relation correctness while discouraging fluent but unsupported operational narratives. For T8 open-ended rejection items, the audit checks whether the response describes visible non-harbor scene elements and avoids unsupported harbor-specific claims such as quay, berth, terminal, docked vessel, cargo handling, or port logistics when those elements are absent.

\subsection{Image-level Correlation and Reporting}
The distributed evaluator records both the number of input items and the number with numeric scores for each track. Closed-form parsing failures are scored as incorrect when an accepted-answer comparison is available; open-ended items require a separate private judging step. The available-track diagnostic returned by the public evaluator is not, by itself, the complete eight-track headline score. Inventory counts and numerically scored counts are distinct quantities and must be reconciled when merging the track results.

HarborEval is item-based because different diagnostic questions can be derived from the same image. This design improves coverage of capability dimensions without requiring a much larger image set, but it introduces image-level correlation. The package has 471 unique images, an average of 2.64 items per image, and a maximum of 7 items per image. Most tracks have approximately one item per image, while T8 has 181 items over 63 unique images, averaging 2.87 items per image. For this reason, the main paper reports item-level metrics for compactness and treats image-level macro scores as an auxiliary analysis.

\subsection{Remaining Benchmark Boundaries}
The audit does not remove all sources of ambiguity. Some harbor functional zones are genuinely mixed, some spatial relations are boundary-dependent, and grid localization can remain ambiguous when objects span multiple cells. The accepted-answer mechanism reduces avoidable unfairness but does not replace visual inspection for borderline cases. Similarly, the T1 and T3 auxiliary yes/no subsets are retained for diagnostic completeness but are not used alone as evidence of scenario understanding. The T8 track is intentionally a rejection diagnostic and should be interpreted as a hallucination-control probe rather than as a general scene-understanding benchmark. These boundaries are consistent with the role of HarborEval in this paper: it is a compact diagnostic protocol for scenario-specific RS multimodal large language model (RS-MLLM) specialization, not an exhaustive benchmark for all harbor RS tasks.

\section{Implementation and Reproducibility Details}
\label{app:implementation_reproducibility}

\begin{table*}[t]

\centering
\caption{Stage-wise training settings for the harbor experiments. LR denotes learning rate; step counts and effective batch size are approximate.}
\label{tab:training_configuration}
\small
\setlength{\tabcolsep}{4pt}
\renewcommand{\arraystretch}{1.12}
\begin{tabularx}{\textwidth}{@{}>{\raggedright\arraybackslash}p{0.11\textwidth}>{\raggedright\arraybackslash}p{0.20\textwidth}>{\raggedright\arraybackslash}p{0.21\textwidth}>{\raggedright\arraybackslash}X>{\raggedright\arraybackslash}X@{}}
\toprule
\rowcolor[HTML]{F2F2F2}
\multicolumn{5}{l}{\textbf{Stage-specific settings}} \\
\midrule
\rowcolor[HTML]{F2F2F2}
\textbf{Stage} & \textbf{Supervision} & \textbf{Exposure} & \textbf{LLaVA-v1.5} & \textbf{Qwen3-VL} \\
\midrule
$S_1$ & RS-Anchor: 569,853 RGB pairs & One epoch; $\sim$35.6K steps & Principally projector; LR $1\times10^{-3}$ & LoRA; LR $1\times10^{-4}$ \\
$S_2$ & Bridging-Conv: 187,296 records & One epoch; $\sim$11.7K steps & Projector LR $2\times10^{-4}$; LoRA LR $1\times10^{-4}$ & LoRA LR $1\times10^{-4}$
 \\
$S_3$ & Scenario-EG: 53,000 records & Two epochs\textsuperscript{a}; $\sim$3.3K steps/epoch & LoRA LR $1\times10^{-4}$ & LoRA LR $3\times10^{-5}$ \\
\bottomrule
\end{tabularx}
\par\vspace{6pt}
\begin{tabularx}{\textwidth}{@{}>{\raggedright\arraybackslash}X>{\centering\arraybackslash}p{0.25\textwidth}>{\centering\arraybackslash}p{0.25\textwidth}@{}}
\toprule
\rowcolor[HTML]{F2F2F2}
\multicolumn{3}{l}{\textbf{Backbone-specific settings}} \\
\midrule
\rowcolor[HTML]{F2F2F2}
\textbf{Setting} & \textbf{LLaVA-v1.5} & \textbf{Qwen3-VL} \\
\midrule
Maximum sequence length & 2048; S2: 4096 & 2048 \\
Per-device batch size & S1: 4; S2--S3: 2 & 1--2 \\
Gradient accumulation & S1: 4; S2--S3: 8 & 8--16 \\
Effective batch size & $\sim$16 & $\sim$16 \\
\bottomrule
\end{tabularx}
\par\vspace{3pt}
\begin{minipage}{\textwidth}\footnotesize
\textsuperscript{a}FBA $S_3$ and Direct-SFT on $D_3$ use two epochs on both backbones. The collapsed routes also use two epochs of $D_3$ exposure; S1+Collapsed additionally uses one epoch of $S_1$. Qwen3-VL uses an image-token budget of 512. Shared optimizer and adapter settings are described in the text.
\end{minipage}
\end{table*}

Unless otherwise specified by a frozen intermediate-stage manifest, LoRA targets \texttt{q\_proj}, \texttt{k\_proj}, \texttt{v\_proj}, and \texttt{o\_proj}, with rank 64, alpha 128, and dropout 0.05. Selected Stage~2 intermediate adapters use rank 32 and alpha 64; terminal reported routes use rank 64 and alpha 128. The SFT setup uses AdamW with $\beta=(0.9,0.95)$ and $\epsilon=10^{-8}$, cosine scheduling, warmup ratio 0.03, maximum gradient norm 1.0, and bf16 precision. Weight decay is run-specific: the archived LLaVA projector-alignment configuration uses 0, the Direct-SFT configurations use 0.03, and the staged SFT configurations use 0.05.

The backbone definitions, stage-wise exposure, optimizer and adapter settings, checkpoint policy, deterministic inference, and hardware are given in main-paper Section~\ref{sec:experimental_setup} and Table~\ref{tab:training_configuration}.

\paragraph[Lightweight urban training protocol]{Lightweight urban training protocol}
The urban experiments in main-paper Section~\ref{sec:urban_extension} use Qwen3-VL-2B-Instruct and the supervision pools specified in main-paper Table~\ref{tab:urban_supervision_pools}. All urban training phases use one epoch. The collapsed phase shuffles $D_2\cup D_3$ and processes each sample once. FBA and $B_0\rightarrow S_1\rightarrow C_{2,3}$ therefore use the same three pools with matched sample exposure.

The urban routes continue the LoRA adapter from the preceding training phase, with the optimizer and learning-rate scheduler reinitialized at each transition. Stage-specific learning rates follow the supervision role across order permutations. Evaluation uses the EarthVQA Urban Val, VRSBench-VQA, and RSVQA-HR subsets described in the main paper, with their scores reported separately.

\paragraph{Additional run-manifest controls}
S1+Collapsed loads the Stage~1 checkpoint before the collapsed $D_2\cup D_3$ SFT exposure. Stage~2 modality proportions are RGB 52.9\%, SAR 16.0\%, NIR 15.2\%, and PAN 15.9\%. Stage~3 proportions are RGB 70.1\%, SAR 23.7\%, NIR 2.2\%, and PAN 4.0\%, with about 15\% non-harbor or ambiguous hard negatives. Training follows predeclared exposure rather than checkpoint selection by a validation score. Evaluation records and answer-bearing fields remain outside training manifests. Private answers, accepted alternatives, evidence annotations, scoring rubrics, verifier rationales, teacher metadata, and source-side construction notes are excluded from public inference inputs.

\paragraph{Recovery and audit trail}
Intermediate \texttt{checkpoint-*} states are retained only for recovery and audit; the reported model is normally \texttt{checkpoint-final} after the planned epochs. Wall-clock duration varies substantially by backbone and stage and is not treated as a defining experimental condition. A complete audit requires the mapping from each reported run to its training configuration, final checkpoint, evaluation inputs, raw predictions, normalized outputs, dimension-level scores, and expert-scoring sheets. These records are not all included in the distributed review artifact. Evaluation records and answer-bearing fields remain outside training manifests.

\section{Evaluation Protocol and Scoring Details}
\label{app:evaluation_protocol_scoring}

Comparisons use the same benchmark items and scoring criteria within each evaluation set. The controlled LLaVA-v1.5 and Qwen3-VL HarborEval evaluators share the task-prompt function, with backbone-specific input formatting. External RS-MLLMs use their recommended prompts when available, so identical prompt wording is not assumed across the external-model comparison. HarborEval exposes only the item identifier, image reference, question, answer choices when applicable, question type, and track identifier used for evaluator-side grouping. Answer keys, accepted alternatives, reference answers, evidence annotations, modality labels, audit notes, and scoring rubrics remain private and are merged only after inference. For reproducibility, the evaluation record retains raw predictions, normalized predictions, closed-form and open-ended scores, hallucination flags, dimension-level assessments, and adjudication notes. This separation keeps the scoring process auditable while focusing the paper description on evaluator-visible logic.

\subsection{Prediction Normalization}
The normalizer is used to remove superficial formatting loss rather than to correct semantic errors. It extracts option keys from variants such as ``A'', ``A.'', or ``Option A'', maps option text back to keys when the selected option is unambiguous, standardizes whitespace and Unicode variants, maps yes/no/unknown aliases to the canonical decision set, maps grid synonyms such as ``upper-left'' to \textit{top-left}, and extracts explicit T5/T6 decisions from short explanatory answers. Each normalization action is logged so that raw and normalized scores can be compared. A large raw-normalized gap is treated as a format-sensitivity diagnostic, while the normalized score is used as the main reported score.

\subsection{Auxiliary Closed-form Diagnostics}
HarborEval uses track-specific metrics normalized to $[0,1]$. Multiple-choice and yes/no/unknown items are scored by accepted-answer accuracy:
\begin{equation}
q_i = \mathbf{1}[\hat{a}_i \in \mathcal{A}_i],
\end{equation}
where $q_i\in\{0,1\}$ is the item-level correctness score, $\hat{a}_i$ is the normalized prediction, and $\mathcal{A}_i$ is the accepted answer set; $\mathbf{1}[\cdot]$ is the indicator function. A strict score compares $\hat{a}_i$ with the canonical answer only. Accepted alternatives account for visually ambiguous functional-zone and calibrated-evidence items.

Multi-label T1 object questions are scored by set F1. Given predicted set $\hat{Y}$ and gold set $Y$, precision, recall, and F1 are computed as
\begin{equation}
P=\frac{|\hat{Y}\cap Y|}{|\hat{Y}|},\quad
R=\frac{|\hat{Y}\cap Y|}{|Y|},\quad
F_1=\frac{2PR}{P+R},
\end{equation}
The score is 1 when both sets are empty and 0 when exactly one is empty or the sets have no overlap. Set F1 balances object coverage with prediction precision.

T4 grid grounding uses a $3\times3$ grid with the canonical cells \textit{top-left}, \textit{top-center}, \textit{top-right}, \textit{middle-left}, \textit{middle-center}, \textit{middle-right}, \textit{bottom-left}, \textit{bottom-center}, and \textit{bottom-right}. Because vessels, basins, piers, and storage regions can straddle cell boundaries, the main T4 metric is soft grid F1. Pairwise cell similarity is 1.0 for the same cell, 0.5 for edge-adjacent cells, 0.25 for diagonal-adjacent cells, and 0 otherwise. Each item stores one audited accepted-cell set. The scorer performs deterministic one-to-one matching from high to low similarity; ties are resolved using a fixed cell ordering. Soft precision, soft recall, and soft F1 are defined as
\begin{equation}
\label{eq:soft_grid_f1}
P_g=\frac{m_g}{|\hat{G}|},\qquad R_g=\frac{m_g}{|G|},\qquad F_g=\frac{2P_gR_g}{P_g+R_g}.
\end{equation}

where $m_g$ is the summed soft match score, $\hat{G}$ is the predicted grid-cell set, and $G$ is the accepted grid-cell set. An empty predicted or accepted set gives a soft grid F1 score of 0. The score is also 0 when the summed match score is 0. Exact grid F1, precision, recall, and exact match are retained as auxiliary diagnostics.

T5 and T6 use the same decision vocabulary, \textit{Yes}, \textit{No}, and \textit{Cannot determine}, but probe different evidence roles. T5 asks whether a property is observable under the supplied modality and image quality; T6 asks whether a visually grounded claim is supported, contradicted, or not decidable from the image. The main T5/T6 score is semantic decision accuracy over accepted answers, while strict accuracy compares only with the canonical decision. During the final audit, weakly visual T5/T6 records are rewritten or replaced, including 64 visual replacements, so that these tracks emphasize visible evidence, modality observability, and evidence-based rejection rather than text priors.

We additionally report closed-form micro accuracy, strict variants, T5/T6 semantic and strict accuracies, T4 exact-grid diagnostics, and T7 rubric dimensions to make clear whether a gain comes from structured VQA, spatial grounding, evidence calibration, rejection, or open-ended reporting.

\subsection{Open-ended and Public-source Score Records}
T7 evaluates whether a model can produce a grounded RS report rather than merely select an option. Each T7 scoring packet contains the image, question, reference answer, key visible objects, evidence or relation fields when available, positive criteria, forbidden claims, and an anonymized model response. A fixed image-grounded multimodal judge scores six dimensions: groundedness, object coverage, functional-zone understanding, spatial relation, uncertainty and rejection, and concision. Each dimension is scored on a 0--5 scale, and the aggregate is normalized to the reporting scale. The scoring rubric has higher priority than optional key points: a response is rewarded for accurate visible evidence and penalized for unsupported port names, vessel identities, cargo types, coordinates, operating status, temporal claims, or objects absent from the image.

OpenEval applies the same six reporting criteria through expert review, with forbidden-claim flags recorded separately. Model identities and training routes are hidden from the scoring sheet. The score record keeps the anonymized sample identifier, anonymous model identifier, dimension-level scores, hallucination or forbidden-claim flags, short reviewer notes, and adjudication status when a borderline or inconsistent case requires review by an additional expert.

\paragraph{Public Harbor scoring}
Public Harbor is the collective name for the VRSBench- and RSVQA-derived RGB harbor subsets used to assess harbor-domain generalization~\citep{lobry2020rsvqa,li2024vrsbench}. Their results are reported separately in main-paper Table~\ref{tab:rsmllm_compare}. It contains 580 public test-style samples over 200 images, including 430 VQA items and 150 image-captioning items. The VQA portion covers object category, existence, quantity, color, shape, size, position, direction, scene type, and reasoning questions, while the captioning portion evaluates detailed image description.

Public Harbor is maintained as an evaluation-only partition separate from the stage-wise training pools and from HarborEval. Its reference answers and captions are hidden during inference and merged only by the scorer. This keeps the subset useful as an external public-source generalization check rather than as an additional source of training supervision.

For VQA, we follow the VRSBench semantic matching protocol: answers are first evaluated by relaxed substring matching, yes/no and numerical answers are evaluated by strict exact match, and remaining open-set answers are evaluated by an LLM semantic matcher~\citep{li2024vrsbench}. For captioning, we use a CLAIR-style LLM score measuring semantic consistency between the generated and reference captions~\citep{chan2023clair}. Let $n_{\mathrm{vqa}}$ and $n_{\mathrm{cap}}$ be the numbers of VQA and captioning samples. The final Public Harbor score is computed as a sample-weighted aggregate after mapping both components to a 0--100 scale:
\begin{equation}
\label{eq:public_harbor_score}
S_{\mathrm{PH}}
=
\frac{
n_{\mathrm{vqa}}A_{\mathrm{VQA}}
 + n_{\mathrm{cap}}(100C_{\mathrm{CLAIR}})
}{
n_{\mathrm{vqa}}+n_{\mathrm{cap}}
},
\end{equation}
where $A_{\mathrm{VQA}}$ is VQA accuracy on the 0--100 scale and $C_{\mathrm{CLAIR}}$ is the caption semantic-consistency score on the 0--1 scale. For Public Harbor, $n_{\mathrm{vqa}}=430$ and $n_{\mathrm{cap}}=150$. This expression defines the package-level aggregate; it does not replace the separate VRSBench and RSVQA scores reported in the main results.

\subsection{Stage-validation Scoring}
\label{app:stage_validation}
\paragraph{RS-VL Val}
RS-VL Val. evaluates RGB land-cover discrimination through ten-candidate caption retrieval and length-matched pairwise comparison. Its primary score equally weights retrieval R@1 and pairwise accuracy:
\begin{equation}
\label{eq:rsvl_main}
S_{\mathrm{RSVL}}=\tfrac{1}{2}\bigl(\mathrm{R@}1+\mathrm{Acc}_{\mathrm{pw}}\bigr),
\end{equation}
Here, R@1 is the fraction of retrieval items for which the positive caption ranks first, and $\mathrm{Acc}_{\mathrm{pw}}$ is the fraction of correctly answered pairwise items. Both components and their mean $S_{\mathrm{RSVL}}$ lie in $[0,1]$; the composite is multiplied by 100 for reporting.

\paragraph{MultiSource Val}
The primary MultiSource Val. score evaluates image descriptions generated without a modality label. A fixed image-grounded judge assesses each generated response $i$ in six dimensions: scene understanding ($d_{i1}$), object recognition ($d_{i2}$), spatial--functional reasoning ($d_{i3}$), modality-evidence use ($d_{i4}$), evidence calibration ($d_{i5}$), and clarity ($d_{i6}$), each on a 0--100 scale:
\begin{equation}
\label{eq:ms_gen_composite}
s^{\mathrm{gen}}_{\mathrm{MS},i}=\sum_{j=1}^{6}\omega_jd_{ij},
\end{equation}
Here, $d_{ij}\in[0,100]$ is the score of response $i$ on dimension $j$, and $\omega_j$ is its weight. The weight vector $\boldsymbol{\omega}=(0.20,0.25,0.20,0.20,0.10,0.05)$ follows the dimension order above and sums to one. Response scores are averaged within each modality; the reported $S^{\mathrm{gen}}_{\mathrm{MS}}$ is the unweighted mean of the resulting modality averages over the modalities present in the evaluation. Clarity receives a smaller weight to prioritize visual correctness over fluency. The auxiliary diagnostics are described below.

Both stage-validation sets pair each image with one \emph{retrieval} item and one \emph{pairwise} item. Retrieval uses one positive caption and nine TF--IDF hard negatives; pairwise contrasts the positive with the length-matched hardest negative among the top-$10$ candidates, with A/B order randomized. For positive rank $r$, we report R@$k=\mathbf{1}[r\le k]$, MRR $=1/r$, and NDCG $=1/\log_2(r+1)$, averaged over items. Pairwise accuracy is $\mathrm{Acc}_{\mathrm{pw}}=\frac{1}{N}\sum_i \mathbf{1}[\hat{y}_i=y_i]$, where $N$ is the number of pairwise items, $\hat{y}_i$ is the predicted option, and $y_i$ is the correct option. A-rate bias checks are retained.

The retrieval and pairwise items are paired by image so that one branch does not receive an easier image distribution than the other. Hard negatives are selected from the frozen validation candidate pool rather than from the training pools, and the randomized A/B order is retained with the parsed prediction. We inspect A-rate bias and degenerate single-option behavior before aggregation; malformed or non-parseable answers are kept as errors rather than manually repaired.

\paragraph{Auxiliary diagnostics and dataset scope}
RS-VL Val. uses 356 fixed RGB images. Generative backbones emit a ranking directly, single-LLM backbones use per-candidate answer likelihood, and an auxiliary PMI ranking is reported only when image-conditioned likelihood behaves consistently. This diagnostic is used for within-backbone stage trajectories and does not replace public RS retrieval or captioning benchmarks.

MultiSource Val. contains 320 images across RGB, SAR, PAN, and NIR (approximately 80 per modality) and uses same-sensor hard negatives for its auxiliary discrimination track. The private modality label is available only to the evaluator for enforcing sensor-specific evidence boundaries. Severe scene errors, non-RGB color hallucinations, fabricated identities, and degenerate or refused answers are reflected in the corresponding dimension scores and retained as failure flags. Modality-recognition accuracy is reported separately when the prompt requests a sensor guess. Equal weighting of modality means prevents differences in subset size from dominating the generation diagnostic.

\section{Prompt and Rubric Templates}
\label{app:prompt_rubric_templates}

\paragraph{Expanded reporting criteria}
The criteria in main-paper Table~\ref{tab:appendix_openeval_rubric} retain the same evidence requirements. Object coverage includes relevant visible vessels, docks, basins, storage, roads, buildings, and vegetation. Functional-zone interpretations may be berthing, navigation, storage, logistics, mixed, uncertain, or non-harbor when supported by the image; confident labels for unsupported or ambiguous regions are penalized. Spatial criteria include adjacency, containment, grid location, and object--region relations. When evidence is insufficient, including under modality or resolution limits, \textit{Cannot determine} or explicit rejection is rewarded rather than a forced harbor interpretation. Concision rewards useful, compact, non-redundant reporting.

The common reporting rubric is given in main-paper Table~\ref{tab:appendix_openeval_rubric}, and semantic-prompt controls in main-paper Section~\ref{sec:experimental_setup}. Table~\ref{tab:appendix_prompt_templates} retains the full semi-structured inference templates for HarborEval and Public Harbor.

Prompt normalization follows four practical rules. First, each prompt separates the task instruction from the answer channel, so that closed-form tracks can be parsed from an explicit final answer while open-ended tracks can still ask for concise supporting evidence. Second, the prompt never exposes modality labels, track names, accepted alternatives, evidence objects, or evaluator-side rubrics; these fields are merged only after inference. Third, prompts avoid dataset-specific wording and file-name cues, because the goal is to evaluate visual reasoning rather than benchmark memorization. Finally, the controlled LLaVA-v1.5 and Qwen3-VL comparisons share the same semantic task instructions within backbone-specific wrappers. External RS-MLLMs use their recommended prompts when available.

For closed-form tracks, the scorer reads the first unambiguous final-answer token or option key. Multi-select responses are normalized to option sets and evaluated using the set F1 metric defined in Section~\ref{app:evaluation_protocol_scoring}. Yes/no/unknown questions are intentionally phrased as evidence decisions: \emph{Yes} requires clear positive evidence, \emph{No} requires contradiction or modality-incompatible evidence, and \emph{Cannot determine} is used when the target may be plausible but the image does not support the claim. This convention is important for SAR, PAN, and NIR inputs, where color, identity, cargo, ownership, operator, temporal status, and precise location claims are often not visually recoverable.

For open-ended reporting, the prompt asks for a compact visual report rather than a free-form story. Responses are rewarded for naming visible objects, functional zones, spatial relations, and uncertainty when appropriate, but penalized for unsupported port names, vessel identities, cargo types, operators, exact operations, or overconfident harbor interpretation under weak evidence. The shared evidence criteria provide a common auditing perspective for T7, T8 open reports, Public Harbor captions, and OpenEval despite their different output lengths.

Prompt instances are generated deterministically from the frozen public inference partition. The image, question, and options are inserted before applying the backbone-specific conversation wrapper. Decoding is fixed within each comparison and sampling is disabled for closed-form questions. The parser scores the explicit final token, retains any explanation for audit, and marks an unparseable response as ambiguous rather than repairing it manually. Thus, the semi-structured templates expose the invariant evidence, ambiguity, and output rules without implying that all backbones share an identical literal prompt. Evaluation inputs, model responses, parsed decisions, and score records are retained together so that abnormal rejection and modality results remain traceable to the exact response.

\begin{table*}[t]
\centering
\caption{Semi-structured inference prompt templates used for HarborEval and Public Harbor. Backbone-specific chat wrappers add image tokens or conversation delimiters around the same semantic instructions.}
\label{tab:appendix_prompt_templates}
\begin{tabularx}{\linewidth}{@{}>{\raggedright\arraybackslash}X@{}}
\toprule
\rowcolor[HTML]{F2F2F2}
{\small\textbf{P1 Closed-Form Recognition / Relation / Rejection}}\tabularnewline
{\fontsize{9}{10}\selectfont\ttfamily Goal: answer a structured harbor-diagnostic question from the image only.\newline Tracks: T1 object support; T2 dominant functional zone; T3 relation or geometry; T8 non-harbor rejection.\newline Inputs: \{image\}, \{question\}, \{options when applicable\}.\newline Rules: use visible evidence; ignore filenames, metadata, geography priors, and hidden labels; choose all valid options only when the task is multi-select.\newline Output: FINAL = \{option key(s) | Yes | No | Cannot determine\}.\par}\tabularnewline
\midrule
\rowcolor[HTML]{F2F2F2}
{\small\textbf{P2 Grid Grounding}}\tabularnewline
{\fontsize{9}{10}\selectfont\ttfamily Goal: localize the visually supported target without bounding boxes.\newline Grid: split the full image into a fixed 3x3 layout with standard cell names.\newline Inputs: \{image\}, \{target or question\}.\newline Rules: select every cell visibly occupied by the target; include multiple cells for spanning objects; do not pad with neighboring cells for caution.\newline Output: FINAL = \{comma-separated grid cells\}.\par}\tabularnewline
\midrule
\rowcolor[HTML]{F2F2F2}
{\small\textbf{P3 Evidence Decision and Modality Constraint}}\tabularnewline
{\fontsize{9}{10}\selectfont\ttfamily Goal: decide whether a claim is supported under the image modality.\newline Inputs: \{image\}, \{modality-blind question\}, \{choices when applicable\}.\newline Decision policy: Yes = clear positive evidence; No = clear contradiction or image-incompatible claim; Cannot determine = plausible but insufficient evidence.\newline Forbid: cargo type, vessel identity, operator, ownership, exact location, time, throughput, or operational status unless directly visible.\newline Output: FINAL = \{Yes|No|Cannot determine\}; EVIDENCE = \{one concise visual reason\}.\par}\tabularnewline
\midrule
\rowcolor[HTML]{F2F2F2}
{\small\textbf{P4 Open Grounded Reporting}}\tabularnewline
{\fontsize{9}{10}\selectfont\ttfamily Goal: produce a concise report or short answer grounded in visible evidence.\newline Variants: T7 harbor description; T8 visible-scene report under rejection setting; Public Harbor VQA; Public Harbor captioning.\newline Cover when defensible: major objects, functional zones, spatial relations, modality-compatible cues, uncertainty, and missing evidence.\newline Avoid: named locations, identities, cargo, operators, precise operations, or overconfident port interpretation when evidence is insufficient.\newline Output: DIRECT ANSWER = \{grounded report or concise answer\}.\par}\tabularnewline
\bottomrule
\end{tabularx}

\end{table*}

\section{Qualitative Evidence-Grounding Comparison}
\label{app:principle_visualization}
Figure~\ref{fig:principle_hardneg} compares a general-purpose MLLM with FBA on two harbor-related tasks. The upper response in each case is produced by the general model, and the lower response by FBA. The comparison highlights how scenario specialization changes the use of visual evidence in both harbor recognition and hard-negative rejection.

In the positive case, the question asks for visual evidence supporting a harbor interpretation. The general model gives a broad description of boats in the water. FBA identifies a vessel docked at a pier together with a nearby warehouse and parking areas, organizing the objects and their spatial relations into a scene-level explanation.

In the hard-negative case, the general model predicts a harbor from an image of a single vessel in open water. FBA examines the surrounding scene, identifies the absence of harbor infrastructure, and gives an evidence-grounded rejection. Together, the paired responses illustrate FBA's ability to connect visible objects with harbor-specific context and make task-relevant decisions. They complement the evidence-judgment and non-harbor-rejection results reported in the main paper.

\section{Qualitative Modality Examples}
\label{app:qualitative_modality_examples}
Main-paper Fig.~\ref{fig:qualitative_harbor_examples} and main-paper Section~\ref{sec:qualitative_examples} introduce the illustrative RGB, PAN, SAR, and NIR panels. The panel-level explanations below retain their response-form and evidence-use context.

The RGB case illustrates the most semantically expressive setting. The response identifies a compact industrial harbor, a large vessel along the central quay, smaller vessels, dockside facilities, and storage buildings, and then organizes these visible elements into a combined berthing, service, and storage zone. The answer describes visible structure and functional layout without inferring a named port, cargo category, operator, throughput, or current activity.

The PAN example instead emphasizes geometric localization. With spectral color unavailable, the answer relies on elongated bright structures, their alignment with the dockside, and proximity to a linear quay edge to select the supported grid cell. This is the intended role of the grid-grounding track: the prediction should reflect the spatial support of the target rather than generic confidence that vessels occur somewhere in the image. The example also shows why modality-blind prompts remain meaningful when the accepted evidence is defined geometrically.

The SAR and NIR cases expose two different evidence boundaries. In SAR, bright backscatter clusters distributed along the waterfront support the presence of dockside infrastructure and vessel-like targets beside a dark water surface, but they do not justify optical color or fine-grained appearance claims. In NIR, the response uses the coastal edge, adjacency between facilities and the waterfront, and orientation toward surrounding water to express a spatial relation. The relevant cue is therefore relational organization rather than natural-color appearance. Together, the two panels illustrate how the same harbor concept can be supported by different sensor-compatible observations.

Across all four panels, a successful answer combines a direct decision with a concise sensor-aware rationale.

The examples illustrate response form and evidence use across sensors. They clarify the qualitative criterion applied throughout HarborEval and OpenEval: reward concise claims that can be connected to observable objects, regions, or relations, and penalize unsupported specificity even when the overall scene is plausibly maritime.

The panels use captioning, grid grounding, evidence-based VQA, and relation reasoning to illustrate a common grounding principle across sensor-specific responses. Comparisons of modality difficulty and average performance rely on the benchmark scores and modality-level diagnostics reported in the main paper.